\documentclass[10pt,twocolumn,letterpaper]{article}

\usepackage[final]{cvpr}              % To produce the CAMERA-READY version
\definecolor{cvprblue}{rgb}{0.21,0.49,0.74}
\usepackage[pagebackref,breaklinks,colorlinks,allcolors=cvprblue]{hyperref}
\usepackage{graphicx}
\usepackage{subcaption}
\usepackage{svg}
\usepackage{placeins}
\usepackage[accsupp]{axessibility} % Improves PDF readability for those with disabilities

\def\paperID{43} % *** Enter the Paper ID here
\def\confName{EarthVision}
\def\confYear{2025}

\title{Better Coherence, Better Height: Fusing Physical Models and Deep Learning \\for Forest Height Estimation from Interferometric SAR Data}

\author{Ragini Bal Mahesh\\
German Aerospace Center (DLR)\\
{\tt\small ragini.mahesh@dlr.de}
\and
Ronny~Hänsch\\
German Aerospace Center (DLR)\\
{\tt\small ronny.haensch@dlr.de}
}

\begin{document}
\maketitle
\begin{abstract}
%About 20-30 lines
%Parameter Retrieval such as forest height estimation from Synthetic Aperture Radar (SAR) images focus on traditional physical models in remote sensing community. While these models are data efficient, easy to interpret and enforce physical plausibility they are based on simplifying assumptions and have generalization issues. Alternatively Deep Learning (DL) approaches have been applied to parameter retrieval however they do not take any physics into consideration. To bridge this gap in this paper, we introduce End-to-End training framework for downstream based output optimization with physical priors. Trained Network has two valuable outputs: the downstream task forest height estimate and corresponding optimized parameter. Moreover, it introduces physics information into the neural network via a novel physics inspired training loss that use a pre-trained neural surrogate model that can be easily integrated into the learning framework to impose the physical constraints. Experiments demonstrate that the End-to-End framework estimates enhances the physical plausibility and generate accurate forest height estimates. Furthermore, we show that optimized parameter is meaningful and improves the accuracy of the downstream task: forest height estimation. Code is made available at \textcolor{blue}{https://www.examplevisiontech.com/research/deep-learning-for-sar-imagery}
Estimating forest height from Synthetic Aperture Radar (SAR) images often relies on traditional physical models, which, while interpretable and data-efficient, can struggle with generalization. In contrast, Deep Learning (DL) approaches lack physical insight. To address this, we propose CoHNet - an end-to-end framework that combines the best of both worlds: DL optimized with physics-informed constraints. We leverage a pre-trained neural surrogate model to enforce physical plausibility through a unique training loss. Our experiments show that this approach not only improves forest height estimation accuracy but also produces meaningful features that enhance the reliability of predictions.
% All code and models are available at \url{https://github.com/ragbm/CoHNet}.
%Code is made available at \textcolor{blue}{https://www.examplevisiontech.com/research/deep-learning-for-sar-imagery}
\end{abstract}    
\section{Introduction}
\label{sec:intro}

% Forest height estimation plays an important role in a wide range of ecological, environmental, and conservation applications. It supports advancing the understanding of forest ecosystems and promotes sustainable practices, e.g. via improved accuracy of forest biomass and carbon stock estimates
Estimating forest height plays an important role for a wide range of ecological, environmental, and conservation applications. It enhances understanding of forest ecosystems and promotes sustainable practices by improving the accuracy of forest biomass and carbon stock estimates~\cite{choi2021improving} - an information fundamental for comprehending the vital role of forests in carbon sequestration.
It also aligns with global initiatives like REDD+ (Reducing Emissions from Deforestation and Forest Degradation) that aim to reduce emissions from deforestation and forest degradation~\cite{bayrak2016ten}. 
The importance of forest height is also illustrated by the ESA BIOMASS mission, selected in 2013 as the seventh Earth Explorer mission. BIOMASS will provide estimates of forest biomass and height with full coverage over the tropical areas. % exploiting the penetration capabilities of P-bans. 
%To further support the BIOMASS mission development, especially concerning concept verification and the development of geophysical algorithms ESA funded the AfriSAR campaign.  
Beyond carbon estimation, forest height also emerges as a critical parameter for monitoring ecosystem dynamics. 
Frequent and high-resolution remote sensing measurements facilitate tracking of forest growth, disturbances (such as logging or natural events), and regrowth. Such insights are important for assessing the overall health and resilience of forest ecosystems, providing essential data for informed conservation and management decisions~\cite{senf2022seeing}.
\begin{figure}
    \centering
    \includegraphics[width=1\linewidth]{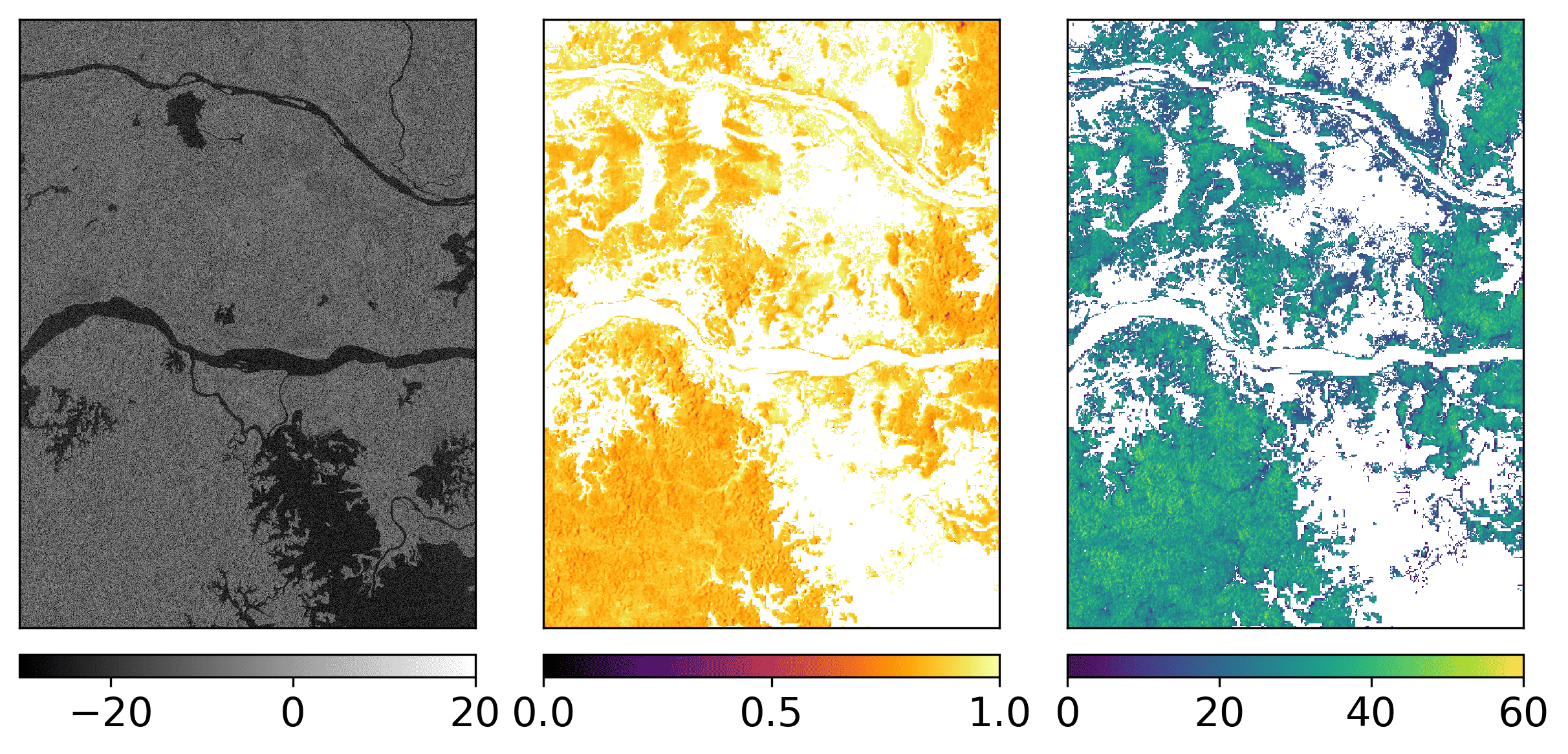}
    \caption{CoHNet, an end-to-end network that uses interferometric coherence estimated from InSAR data (backscatter of one image shown on the left), derives optimized volume decorrelation (center) which is then leveraged by a physical model to estimate forest height (right). White regions in the maps denote non-forest areas. %Extracting true volume decorrelation or Optimized Coherence from radar backscatter. A major scattering contribution that is free from other decorrelations noises
    }
    \label{fig:teaser}
\end{figure}

Synthetic Aperture Radar (SAR) has emerged as one of the most suitable technologies for forest height estimation compared to optical and LiDAR sensors due to several benefits such as robustness regarding weather conditions, day-and-night imaging capabilities, and the ability to partially penetrate the forest canopy. 
SAR systems emit microwave signals towards the Earth's surface and record the backscattered echoes.
The interaction between the radar waves and the forest canopy provides information about the structure and height of the vegetation. 
The conventional approach for forest height estimation using SAR leverages Interferometric SAR (InSAR) data~\cite{guliaev2021forest,kugler2014tandem} and includes establishing an empirical relationship with the interferometric coherence, a correlation variable derived from two SAR images. Coherence can be decomposed %factorized 
into various decorrelation factors, with volume decorrelation being one of the most important contributions~\cite{rizzoli2022derivation}.
This particular factor is intricately linked to the volumetric scattering of the forest canopy. It quantifies the level of decorrelation caused by the three-dimensional nature of the vegetation structure.
Despite its pivotal role, its estimation remains a challenging task. 
Usually a physical model %such as Random Volume over Ground (RVoG) \cite{guliaev2021forest, kugler2014tandem} 
is used in inverting the forest height from the volumetric decorrelation. 
Such models have several limitations in practical applications including potential inaccuracies in representing the complex forest structures, sensitivity to assumptions about the distribution of scatterers, and challenges in handling heterogeneous landscapes.
%In practice, several crucial points affect the computation of the interferometric coherence: First, the SNR value is mostly calculated using a theoretical approach with the assumption of constant noise power over the polarimetric channels \cite{kugler2015forest}. Second, the temporal decorrelation causes a loss of coherence if present. This is not the case in bistatic data from TanDEM-X but temporal decorrelation becomes relevant in repeat pass acquisitions as with Sentinel-1. Third, the window size of the spatial average during coherence computation affects the obtained value. \\

%Several parameters (e.g. the window size of the spatial integration during coherence optimization, compensation terms for quantization terms, etc.) are assumed to be constant over a scene, yet are data-dependent (e.g. large windows would lead to more stable coherence estimates over homogeneous regions but introduce errors in heterogeneous areas). 
Data-driven approaches such as Deep Learning (DL) can perform data-dependent corrections to compensate for modeling errors. 
However, volume decorrelation cannot be directly learned from the data by supervised approaches due to the lack of ground truth data for this variable. 
%Prior work has focused primarily on the advancement of different DL architectures and training methodology \cite{carcereri2023deep} \cite{li2023forest} \cite{mahesh2024forest} \cite{wang2019forest}. 

We propose CoHNet (Coherence and Height Network, %summarized in 
visualization of in- and output maps shown in Figure~\ref{fig:teaser}) as an alternative approach and estimate a DL-optimized parameter that is equivalent to volume decorrelation by optimizing the down-stream task of forest height estimation for which reference data (e.g. via LiDAR measurements) is available and include prior information via a physical model that constraints forest height estimates to physically plausible solutions.
%In this paper, We propose an indirect optimization method to obtain this optimized parameter via an end-to-end deep neural network constrained by a physical model. 
%The a-priori information in the physical model is leveraged to map the input SAR image pair to the target LiDAR-measured forest height while learning the optimized parameter (i.e. volume decorrelation).
In summary, our main contributions include:
\begin{itemize}
    \item We propose CoHNet as an end-to-end framework for downstream-based output optimization with physical priors. The trained network has two outputs: The forest height estimate and optimized volumetric decorrelation.
    \item We introduce physics information into the neural network via a novel physics-inspired training loss that uses a pre-trained surrogate model that can be easily integrated into the learning framework to impose physical constraints.
    %\item We demonstrate through experiments that the End-to-End framework estimates enhance the physical plausibility and generate accurate forest height estimates.  Furthermore, we show that the optimized parameter is meaningful and improves the accuracy of the downstream task: forest height estimation.
\end{itemize}
\section{Related Work}
\label{sec:sota}

%\textbf{Parameter Retrieval Physical Models}: Physical models are essential in interpreting the Interferometric SAR (InSAR) for parameter retrieval such as forest height, as radar signals interact in complex ways with the vegetation. 
\noindent The traditional approach to estimate forest parameters such as height from Interferometric SAR (InSAR) are \textbf{physical models} that
%These models 
simulate how radar waves scatter as they penetrate the canopy, accounting for factors like forest structure and ground surface conditions. 
There are two main categories: 
(i) The Scattering Phase Center Model describes how the radar waves interact with different parts of the forest, such as leaves, branches, and trunks. By modeling the scattering at different canopy layers they provide information on the structure and height \cite{cloude1997polarimetric,huang2021forest,li2023forest,soja2013digital}.
(ii) Interferometric Coherence Models acknowledge that interferometric coherence is affected by vegetation properties such as canopy movement and moisture. These models predict the decorrelation caused by vegetation, which helps to distinguish the canopy from the ground topography within the interferometric signal. 
The most widely used approach is the Random Volume over Ground (RVoG) model \cite{guliaev2021forest} \cite{kugler2014tandem} \cite{papathanassiou2001single} which assumes a vertically homogeneous forest volume over a region. 
Several different extensions have been proposed including the three-stage algorithm \cite{cloude2003three}, the volume temporal decorrelation model \cite{papathanassiou2003effect}, the slope corrected model (S-RVoG) \cite{lu2013s}, and other variations \cite{fu2017combination,zhang2022modified,wang2022evaluation}. 
The RVoG model is based on the assumption that the decay of the forest height along the vertical reflectivity profile is given as a function of the extinction coefficient. %, where the coefficient is assumed to be zero, also the model ignores any ground phase. 
It is simplified to a sinc function which is also called coherence amplitude inversion and can result in overestimation of forest heights due to temporal decorrelation and signal-to-noise decorrelation. 
However, multiple challenges remain, including how signal decorrelation from dense vegetation, topographic complexity, and the influence of moisture influence the radar signals.

Recent \textbf{deep learning based approaches} formulate the task as either a classification or regression problem. Classification frameworks categorize vegetation into discrete height classes, enabling the identification of height ranges corresponding to different forest structures \cite{wang2019forest,yang2023deep} but bearing the risk of not capturing the full range of variation in forest height. Regression-based approaches treat forest height as a continuous value. They are better suited for capturing fine-grained variations in forest height \cite{carcereri2023deep} \cite{li2023forest} \cite{mahesh2024forest} \cite{yang2024catsnet}. 
With the increasing availability of large-scale remote sensing data and the advantage of Deep Learning (DL) in handling complex tasks, many recent studies have focused on fusing data from multiple remote sensing sources such as SAR, LiDAR, and optical data allowing for more robust height estimation \cite{becker2023country,ge2022improved,ge2023deep,li2020high,schwartz2024high}. While supervised learning is the most common approach for training deep learning models using forest height reference data, other learning techniques such as self-supervised learning using masked autoencoders have been used as well \cite{bueso2024forest}. Vision Transformers have also been used to estimate canopy height, e.g. Hy-TeC \cite{Fayad2024} leverages optical and LiDAR data for 10m canopy height mapping over Ghana.

\textbf{Hybrid and physics-aware models} incorporate physics into machine learning models to improve model accuracy, generalization, and reliability, especially in data-sparse domains or applications where parts of the functional relationship between measurements and target variables are known.

One approach is to directly introduce physics into the neural network structure, often referred to as Physics-Informed Neural Networks (PINNs) or physics-aware deep learning - an emerging approach that integrates physical laws and domain-specific knowledge directly into the training of the neural network by embedding a residual loss from a governing equation (e.g. partial differential equations) into the loss function \cite{karniadakis2021physics,mahesh2022physics,raissi2019physics}. For inverse problems where model parameters need to be inferred from data, physical models representing the inversion can be embedded into the loss function and can be used to calculate gradients, helping the network to learn physically consistent parameter values. These models are often hard to differentiate and are thus replaced by surrogate models that are trained by physical model simulation \cite{camps2020living,camps2021physics}.
 
Another approach is to couple physical models with data-driven components like neural networks that for instance can generate more meaningful features. 
Mansour et al. \cite{mansour2024hybrid} propose a hybrid inversion framework to estimate forest height from InSAR coherence where a parametrization of the forest structure is derived as a function of the input features using an MLP. The predicted profile is used to invert the forest height, where the error is backpropagated through the hybrid model components.

We use a physics-aware approach allowing the neural network to learn an optimized parameter relying on the physics-based downstream task which has been encoded in a neural surrogate model.

\section{Methodology}

\subsection{Preliminaries}
\label{sec:prelim}
Interferometric Synthetic Aperture Radar (InSAR) is a remote sensing technique that combines two SAR images taken from slightly different positions enabling measurement of surface height and deformation \cite{moreira2013tutorial}. The key observable of InSAR is the interferometric coherence $\gamma$ between two interferometric images $s_1 (\Vec{w})$ and $s_2 (\Vec{w})$ acquired at given polarization $\Vec{w}$ \cite{carcereri2023deep,rosen2000synthetic,sica2021net,mansour2024hybrid}, i.e.
\begin{equation}
    {\gamma} \left( \kappa_z, \vec{\omega} \right) = \frac{\langle s_1 \left( \vec{\omega} \right) s_2^* \left( \vec{\omega} \right) \rangle}{\sqrt{\langle s_1 \left( \vec{\omega} \right) s_1^* \left( \vec{\omega} \right) \rangle} \sqrt{\langle s_2 \left( \vec{\omega} \right) s_2^* \left( \vec{\omega} \right) \rangle}},
    \label{eq:coh}
\end{equation}
where $(.)^*$ denotes the complex conjugate and $\langle.\rangle$ a spatial average using a predefined window size.
The vertical wavenumber $\kappa_z$ (rad/m) is the sensitivity of the interferometric phase height to changes and is proportional to the baseline distance between the SAR sensors \cite{kugler2014tandem,mansour2024correction,mansour2024hybrid}.

%\begin{equation}
%    \kappa_z = \frac{4 \pi}{\lambda} \frac{\Delta \theta_i}{\sin \theta_i}
%\end{equation}
%where $\lambda$ is the wavelength, $\theta_i$ is the incidence angle, $\Delta \theta_i$ is the difference in the incidence angle due to the spatial baseline or distance between the two image acquisitions, and  factor $m$ depends on the acquisition mode: m = 2 for mono-static mode and m = 1 for bistatic mode.\\ 
Interferometric coherence is commonly derived using a maximum likelihood estimator that employs a moving boxcar window \cite{sica2021net} and can be further factorized \cite{rizzoli2022derivation} into
%\begin{equation}
   $\gamma = \gamma_\text{sensor} \cdot \gamma_\text{temp} \cdot \gamma_\text{quant} \cdot \gamma_\text{SNR} \cdot \gamma_\text{vol},$
   %\label{eq:volume decorrelation}
%\end{equation}
where volume decorrelation~$\gamma_\text{vol}$ is caused by the presence of volumetric scattering and thus of particular relevance for forest height estimation~\cite{guliaev2021forest,kugler2014tandem}. 
It is derived from $\gamma$ by neglecting other decorrelation factors such as quantization decorrelation ($\gamma_\text{quant})$, temporal decorrelation ($\gamma_\text{temp} = \textcolor{black}{1}$) in the case of a bistatic configuration, and decorrelation due to other sensor parameters ($\gamma_\text{sensor} = \gamma_\text{amb} \cdot \gamma_\text{rg} \cdot \gamma_\text{az} $) such as SAR ($\gamma_\text{amb}$) and azimuth ($\gamma_\text{az}$) ambiguities as they do not depend on the illuminated scene. However, $\gamma$ is dependent on the baseline unless proper range filtering actions are made resulting in range decorrelation ($\gamma_\text{rg}$). Thus, usually only a rough approximation of $\gamma_\text{vol}$ is computed as
\begin{equation}
    \gamma_\text{vol} = \gamma\cdot \gamma_\text{rg}/ \gamma_\text{SNR}
    \label{eq:gammavol}
\end{equation}
compensating only for noise ($\gamma_\text{SNR}$) and $\gamma_\text{rg}$.

% \begin{figure}[t]
%     \centering
%     \includegraphics[width=1\linewidth]{Images/RVoG_model.png}
%     \caption{The RVoG model describes the forest structure as a two-layer scattering model with ground elevation~$z_0$ and volume height~$h_v$. $F(z)$ is the radar reflectivity of the forest scatters at different heights~$z$ and decays as a function of the extinction coefficient $\sigma$, the ground phase~$\phi_0$m and the ground-to-volume amplitude ratio~$\mu$ \cite{qi2019improved}.}
%     \label{fig:rvog}
% \end{figure}

The inversion of bio-/geophysical parameters such as forest height includes establishing a relation between the estimated complex coherence and the bio-/geophysical parameter. 
One of the most commonly used approaches for forest height is the Random Volume over Ground (RVoG) model which describes the forest canopy via a two-layer scattering model consisting of the vegetation layer and the ground surface layer. % as shown in Figure~\ref{fig:rvog}. 
The volume scatterers follow a uniform random spatial distribution \cite{qi2019improved,zhang2022modified}. The volume decorrelation factor $\gamma_{vol}$ (estimated via Eq.~\ref{eq:gammavol}) is set into relation to the forest canopy height $h_v$ and the forest vertical reflectivity profile $f(z)$ (where $z$ is forest scatters at different heights and $z_0$ the ground contribution) \cite{brolly2016lidar,mansour2024hybrid} via

\begin{equation}
    {\gamma}_{\text{Vol}} \left( \kappa_z \right) = e^{i \kappa_z z_0} \frac{\int_{0}^{h_v} f(z) \, e^{i \kappa_z z} \, dz}{\int_{0}^{h_v} f(z) \, dz}.
\end{equation}

In practice, several crucial points affect the computation of the interferometric coherence and therefore the estimation of forest height: First, the SNR value is mostly calculated using a theoretical approach with the assumption of a constant noise power \cite{kugler2015forest}. Second, the temporal decorrelation causes a loss of coherence if present. This is not the case in bistatic data from TanDEM-X but temporal decorrelation becomes relevant in repeat pass acquisitions as with Sentinel-1. Third, the window size of the spatial average during coherence computation affects the obtained value. Several of these parameters (e.g. the window size of the spatial integration during coherence optimization, compensation terms for quantization errors, etc.) are assumed to be constant over a scene, yet are actually data-dependent (e.g. large windows would lead to more stable coherence estimates over homogeneous regions but introduce errors in heterogeneous areas). 

% \subsection{Physics-aware Model Architecture}
% \begin{figure*}[t]
%     \centering
%     \includegraphics[width=1\linewidth]{Images/End-to-End-Network.png}
%     \caption{Overview of the proposed framework for forest height estimation. The first network estimates the optimized volume decorrelation $\hat\gamma_\text{Vol}$ from the given coherence $\gamma$. The loss calculated from the output of the second network (with fixed, pre-trained weights) using this parameter together with the vertical wavenumber $\kappa_z$ compared to a reference forest height is back-propagated through the entire pipeline to update the weights of the first network.}
%     \label{fig:model_architecture}
% \end{figure*}

\subsection{CoHNet: A Physics-aware Model Architecture}
\begin{figure*}[t]
    \centering
    % \includesvg[width=0.90\textwidth]{Images/End_to_End_Model_V8_drawio_16_corr.svg}
    \includegraphics[width=0.85\textwidth]{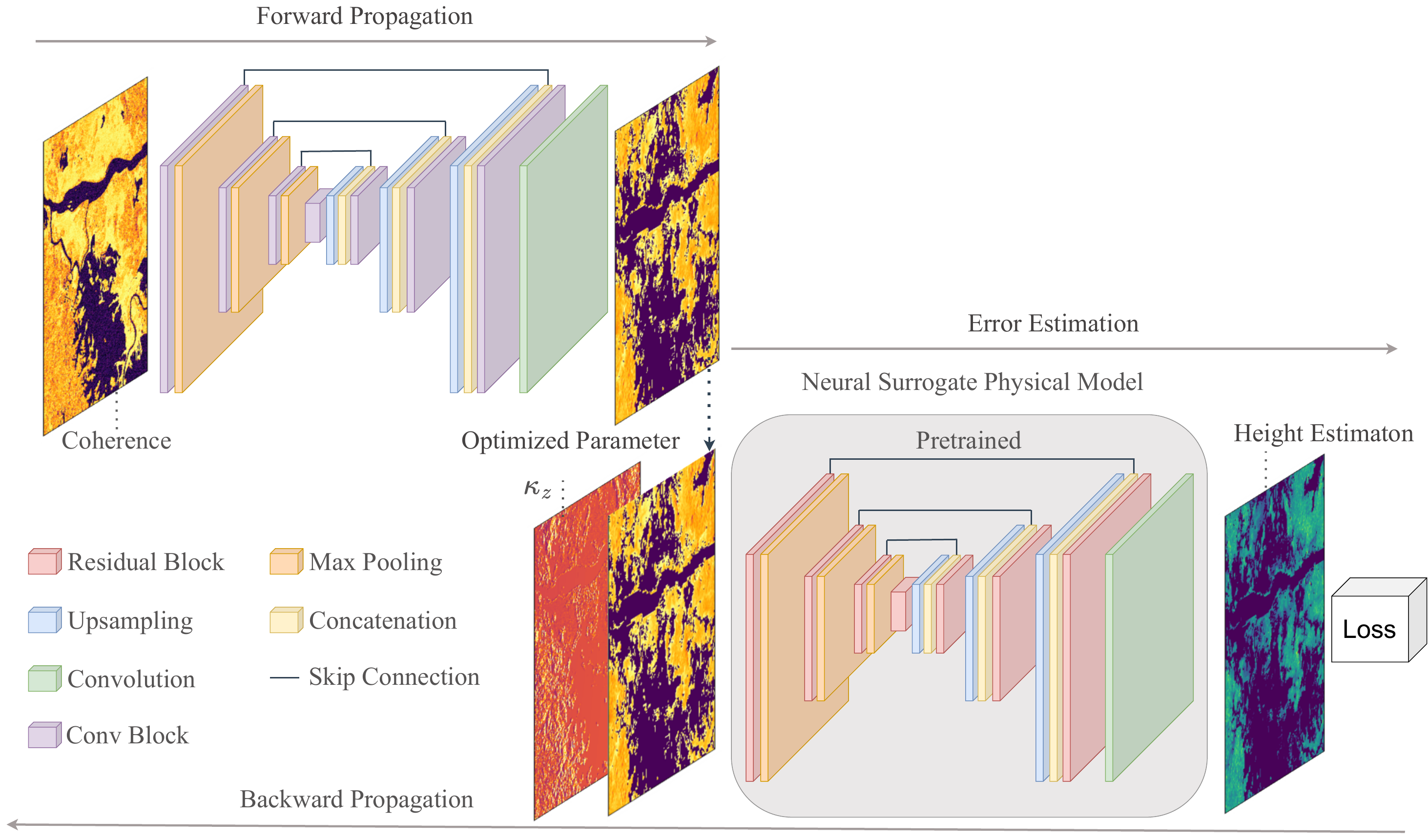}
    \caption{Overview of CoHNet for forest height estimation. The first network estimates the optimized volume decorrelation $\hat\gamma_\text{Vol}$ from the given coherence $\gamma$. The loss calculated from the output of the second network (with fixed, pre-trained weights) using this parameter together with the vertical wavenumber $\kappa_z$ compared to a reference forest height is back-propagated through the entire pipeline to update the weights of the first network.}
    \label{fig:model_architecture}
\end{figure*}

The underlying idea of the proposed framework is that the physical model should in principle be able to provide accurate forest height if its model assumptions are fulfilled. Among others, these include that the presented input is the actual volume decorrelation. 
The commonly used approximation of the volume decorrelation (see Eq.~\ref{eq:gammavol}) is close enough to lead to reasonable results, yet the rather crude assumptions (as discussed in Section~\ref{sec:prelim}) cause estimation errors.
Thus, we propose a network that generates an optimal input for the physical model, minimizing prediction errors and producing estimates as close as possible to the actual forest height.
%Thus, we propose a network which computes an input for the physical model which is optimal in the sense that it leads to the smallest error regarding its predictions, i.e. estimates as close as possible to the actual forest height.

%\begin{figure*}[t]
%    \centering
%    \includegraphics[width=1\linewidth]{Images/End-to-End-Network.png}
%    \caption{Overview of the proposed framework for forest height estimation. The first network estimates the optimized volume decorrelation $\hat\gamma_\text{Vol}$ from the given coherence $\gamma$. The loss calculated from the output of the second network (with fixed, pre-trained weights) using this parameter together with the vertical wavenumber $\kappa_z$ compared to a reference forest height is back-propagated through the entire pipeline to update the weights of the first network.}
%    \label{fig:model_architecture}
%\end{figure*}

Figure~\ref{fig:model_architecture} provides an overview of CoHNet. 
Given the computed coherence as input, it estimates the optimized volumetric decorrelation which is then used in a physical model (together with $\kappa_z$) to predict forest height. 

Unfortunately, the vanilla implementation of the RVoG model does not allow its direct inclusion into such a pipeline as it is neither fast enough during inference nor differentiable. 
We replace it with a Neural network based Surrogate Model (NSM) which is pretrained to simulate the physical model outputs given the corresponding inputs of volume decorrelation and $\kappa_z$.

Both models are concatenated into a single pipeline such that the output of the coherence model is used as input to the forest height model. End-to-end training is performed on the concatenated model by fixing the weights of the pre-trained forest height model (i.e. NSM). Only the coherence model weights are updated using the loss generated by NSM. 
Thus, the final network learns input-output mappings such that the model’s solution space is constrained by the physical model estimates allowing the solution to be in the functional space of the physical model. Once the model is trained, it provides two outputs: The optimized volume decorrelation and forest height estimates.

The first network follows a U-Net architecture proposed for directly estimating forest height from coherence or other InSAR related features \cite{mahesh2024forest} with the exception that the activation function in the output layer is a sigmoid function so that the output is limited to the range of coherence.
The second network is based on $\phi$-Net \cite{sica2021net}, a residual U-Net architecture tailored for InSAR data. We use a U-Net architecture for its effectiveness in extracting geospatial features and its compatibility with the study’s physics constraints. 
%\textcolor{red}{The choice of U-net architecture for our model is due to its effectiveness in extracting geospatial features. While other advanced architectures such as transformers have shown promise in remote sensing applications it may not provide significant advantage over convolutional architectures. Additionally to maintain interoperability with the physics constraints used in this study U-net seems to be a reliable choice.}

%The first network is a U-Net \cite{ronneberger2015u} consisting of an input block and a sequence of four encoder blocks, four decoder blocks, and an output block \cite{mahesh2024forest}. The encode-decoder structure is connected with a convolution block in the bottleneck layer. The convolution block is modified to have two instead of a single double convolution layers. Each convolution layer is followed batch normalization layer and a ReLU activation function layer. A 2D Max pooling layer was used in the encoder block. In addition, layers were added to both the encoder and decoder blocks. The activation function on the output layer is a sigmoid function so that the output is limited to max value of 1 to match the coherence range [0,1].

%\subsection{Physics-Inspired Training Loss}
For the optimization of the network, we introduce a novel physics-inspired loss based on the Root Mean Squared Error between predicted $H^\text{est}$ and reference $H^\text{ref}$ forest heights of $N$ samples, i.e.
\begin{equation}
    \mathcal{L} = \sum_{i=1}^N \sqrt{\frac{\left(H^\text{ref}_i - h_\text{NSM}(\hat\gamma_{Vol}(\gamma_i))\right)^2}{N}},
\end{equation}
where $h_\text{NSM}(\hat\gamma_{Vol}(\gamma_i)) = H^{est}$ is the forest height predicted by the surrogate network $h_\text{NSM}$ of the physical model based on the output of the first network i.e. the optimized volume decorrelation~$\hat\gamma_{Vol}$ based on the given coherence~$\gamma$.  
This allows us to learn optimized volume decorrelation without direct reference data as this does not exist.
We preferred this loss over an L1-based loss since we do not expect outliers within the data and the estimation of taller trees is more relevant than the estimation of lower vegetation.
%The network is optimized via an Adam optimizer. 

% We use a standard Adam optimizer together with a L2 regularization on the network weights for optimization.

%\textcolor{red}{We use the standard L2 loss function for regression task to estimate fixed range of forest heights and no outliers.The network weights are optimized with Adam Optimizer}
\section{Experiments}
\begin{figure*}[ht]
    \centering
    \begin{subfigure}[b]{0.23\textwidth}
        \centering
        \includegraphics[width=1\linewidth]{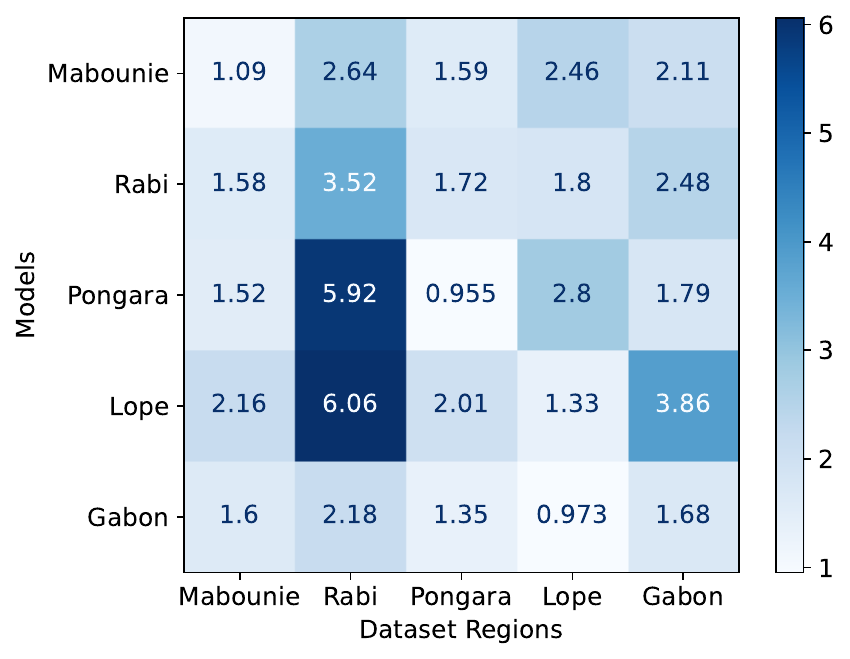}
        \caption{Pre-trained NSM}
        \label{fig:conf_matrix_physics_model}
    \end{subfigure}%
    \hspace{0.01\textwidth}
    \begin{subfigure}{0.23\textwidth}
        \centering
        \includegraphics[width=1\linewidth]{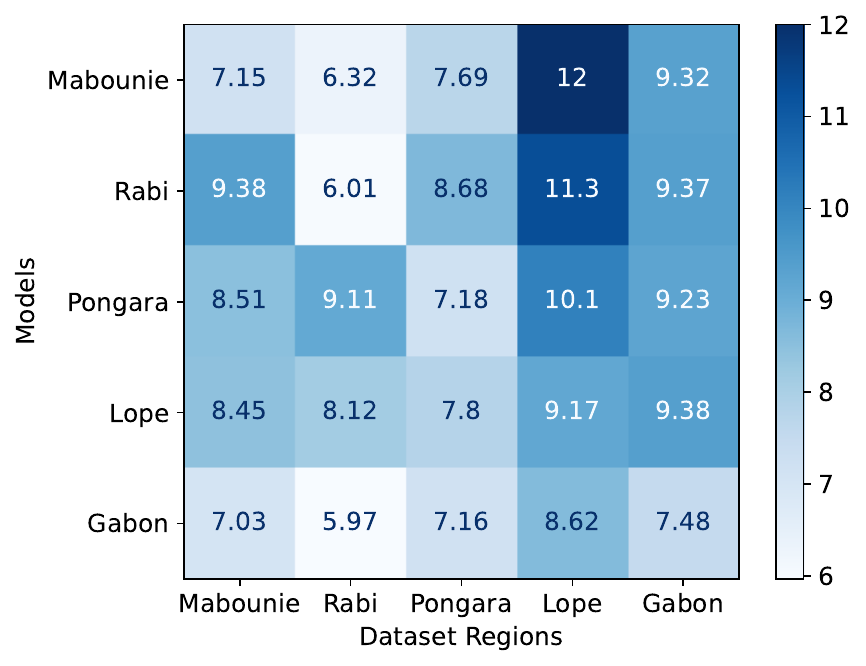}
        \caption{CoHNet with using Gabon NSM}
        \label{fig:conf_matrix_End_to_End_Model_Gabon_PM}
    \end{subfigure}
    \hspace{0.01\textwidth}
    \begin{subfigure}{0.23\textwidth}
        \centering
        \includegraphics[width=1\linewidth]{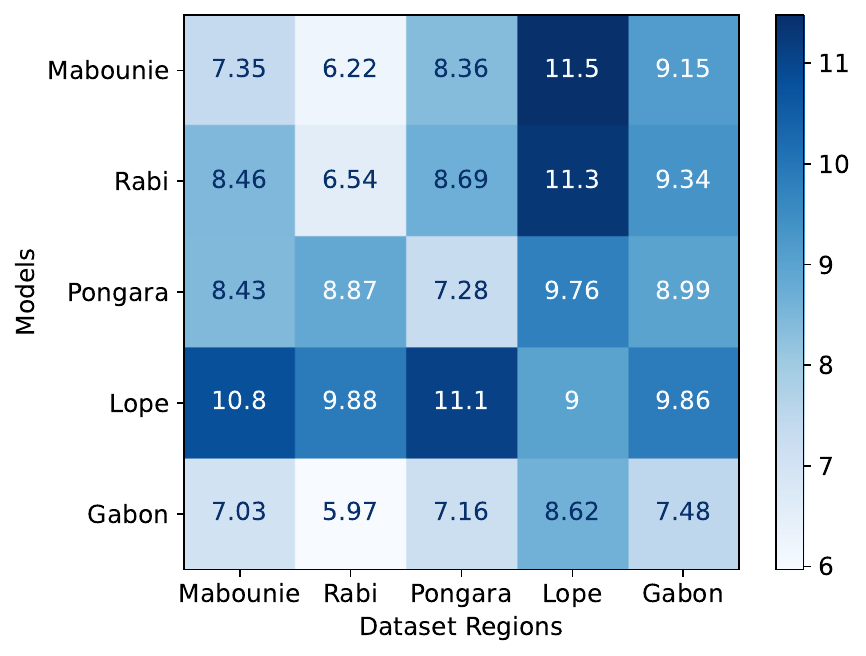}
        \caption{CoHNet with region-wise NSM}
        \label{fig:conf_matrix_End_to_End_Model}
    \end{subfigure}%
    \hspace{0.01\textwidth}
    \begin{subfigure}{0.23\textwidth}
        \centering
        \includegraphics[width= 1\linewidth]{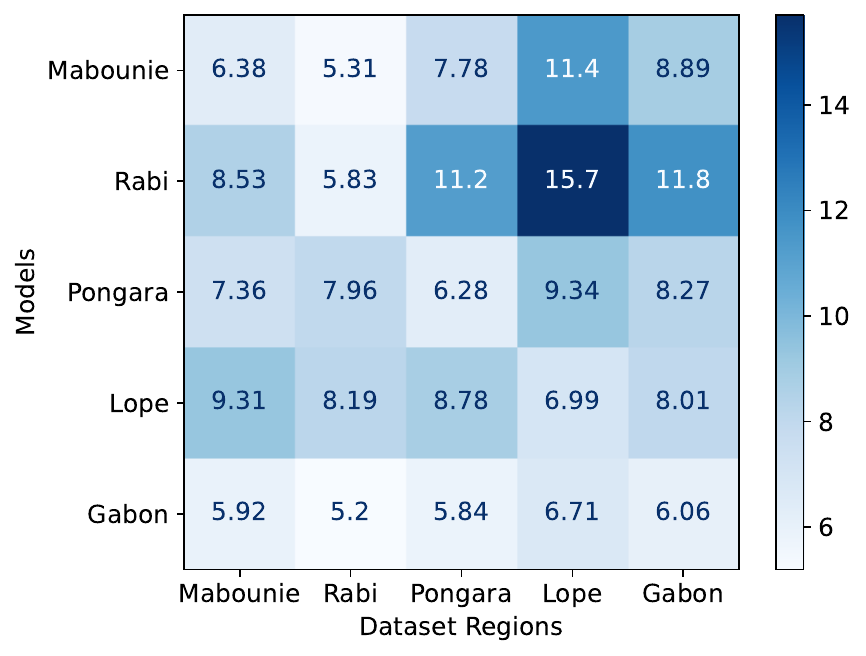}
        \caption{Direct Model.}
        \label{fig:conf_matrix_direct_model}
    \end{subfigure}
    % \caption{Performance of various models trained (rows) and evaluated (columns) over different regions.}
    \caption{Performance (RMSE in [m]) of different models trained (rows) and tested (columns) over various regions. The scale of each heatmap is normalized per subplot for better comparison. Subplots represent (a) Pre-trained NSM, (b) CoHNet with Gabon NSM, (c) CoHNet with region-wise NSM, and (d) Direct Model. Each model demonstrates 
    varying degrees of adaptability %and effectiveness 
    across dataset regions.}
    \label{fig:results}
\end{figure*}

\subsection{Dataset, Setup, and Metrics} 
\label{sec:data}
We train and evaluate CoHNet on data of the AfriSAR campaign \cite{fatoyinbo2021nasa} - a NASA-ESA joint campaign that collected airborne SAR and LiDAR data for sensor calibration and the development of forest structure and biomass retrieval algorithms. 
Instead of the airborne SAR imagery of the AfriSAR campaign, we use InSAR acquisitions obtained by TanDEM-X (DLR) in 2016 over the same region.

The study area for the AfriSAR campaign is Gabon as it is a densely forested country with rich structural and functional biodiversity. 
Five different sites are part of the campaign, i.e. Mondah forest, Lope National Park, Mabounie, Rabi, and Pongara National Park. 
We do not consider Mondah as it is covered by only a single TanDEM-X image which does not allow for unbiased testing.
While this dataset contains only data over Gabon, if offers four geographical disjoint regions with very distinct properties regarding forest type (e.g. Mangrove in Forest in Pongara).
We use the already gridded forest characterization products derived from full-waveform lidar data acquired by NASA's airborne Land, Vegetation and Ice Sensor (LVIS) instrument. 
This contains geolocated canopy cover at a spatial resolution of 25m readily processed to the Level 3 (l3a1) product RH98 which is the mean relative height at which 98\% waveform energy occurs. 

For each of the four regions, the available images are split into training and test images, resulting in 5, 4, 2, and 1 training images for Mabounie, Rabi, Pongara, and Lope, respectively. For each region, we use one additional image as a test set which fits the general data distribution of the training set.
In addition to these four region-wise datasets, a fifth dataset is created that consists of all the four regions combined with a total of 12 training and 4 test images.
Maps of coherence (computed according to Equation~\ref{eq:coh}), volume decorrelation (Equation~\ref{eq:gammavol}), and $\kappa_z$ are geocoded and used by a RVoG model (see Section~\ref{sec:prelim}) to estimate forest height. The average error of the RVoG model over all regions is 14.2~m.
As reference data serves the LVIS measured forest height.
All data products are geocoded, aligned, and divided into $64\times 64$ patches with a stride of 32. 
Table~\ref{tab:stats} summarizes these properties of the train and test sets.

\begin{table}[t]
    \centering
\begin{tabular}{ |c||c|c|c|c|c| }
 \hline
 & \multicolumn{2}{|c|}{\#Patches} & \multicolumn{2}{|c|}{Avg. Height} & RMSE PM \\
 Region & Train & Test & Train & test & (in m) \\
 \hline
 Mabounie & 2985 & 603 & 32.7 & 29.3 & 16.1 \\
 Rabi & 2670 & 621 & 30.5 & 33.6 & 13.3 \\
 Pongara & 1421 & 475 & 31.1 & 29.5 & 14.6 \\
 Lope & 1253 & 848 & 37.8 & 36.4 & 16.8\\ \hline
 Gabon & 8329 & 2547 & 33.0 & 32.2 & 14.2 \\ \hline
\end{tabular}
\caption{Statistics of the used datasets to train and evaluate all models including the number of patches, average LVIS reference height (in m), and the error of the physical model (PM) for each individual region as well as for all regions combined.}
\label{tab:stats}
\end{table}

Training is performed with an Adam optimizer for 100 epochs with a batch size of 8 and a learning rate varying between $10^{-4}$ and $10^{-3}$ 
%Evaluation uses the Root Mean Squared Error (RMSE) between predicted and reference (LVIS) forest height. 
and requires approximately 1 hour on an NVIDIA A100-SXM4-40GB GPU, while inference takes only a few seconds per scene, making the approach computationally feasible for large-scale applications.

\subsection{Surrogate Model Performance}
\label{sec:nsm}
While a physics engine could be used to enforce the physical plausibility, these models are not differentiable, computational expensive, and difficult to integrate into a neural network pipeline.
Thus, we train a Neural network based Surrogate Model (NSM) for the RVoG physical model to leverage physical principles and capture the relationship between the radar backscatter and forest height. The surrogate model is trained with the same inputs as the RVoG model, i.e. volume decorrelation obtained from coherence and the corresponding vertical wavenumber ($\kappa_z$) based on the relative position of the two SAR satellites.

Figure~\ref{fig:conf_matrix_physics_model} shows an overall good agreement between the predictions of the surrogate models and the results of the physical model for all the regions, including the combined Gabon model. 
For surrogate models trained and evaluated on the same region, the RMSE varies between 0.955 to 3.52~m.
The combined Gabon model trained and evaluated on all regions has an RMSE between 0.973 to 2.18~m. 
Surrogate models that are evaluated on regions that they are not trained on show in general still acceptable error rates, yet with a much higher variance.
This shows that replacing the original RoVG model with a neural network based surrogate model is indeed a viable option.

\subsection{CoHNet Performance}
\begin{figure*}[t]
    \centering
    \includegraphics[width=1\textwidth]{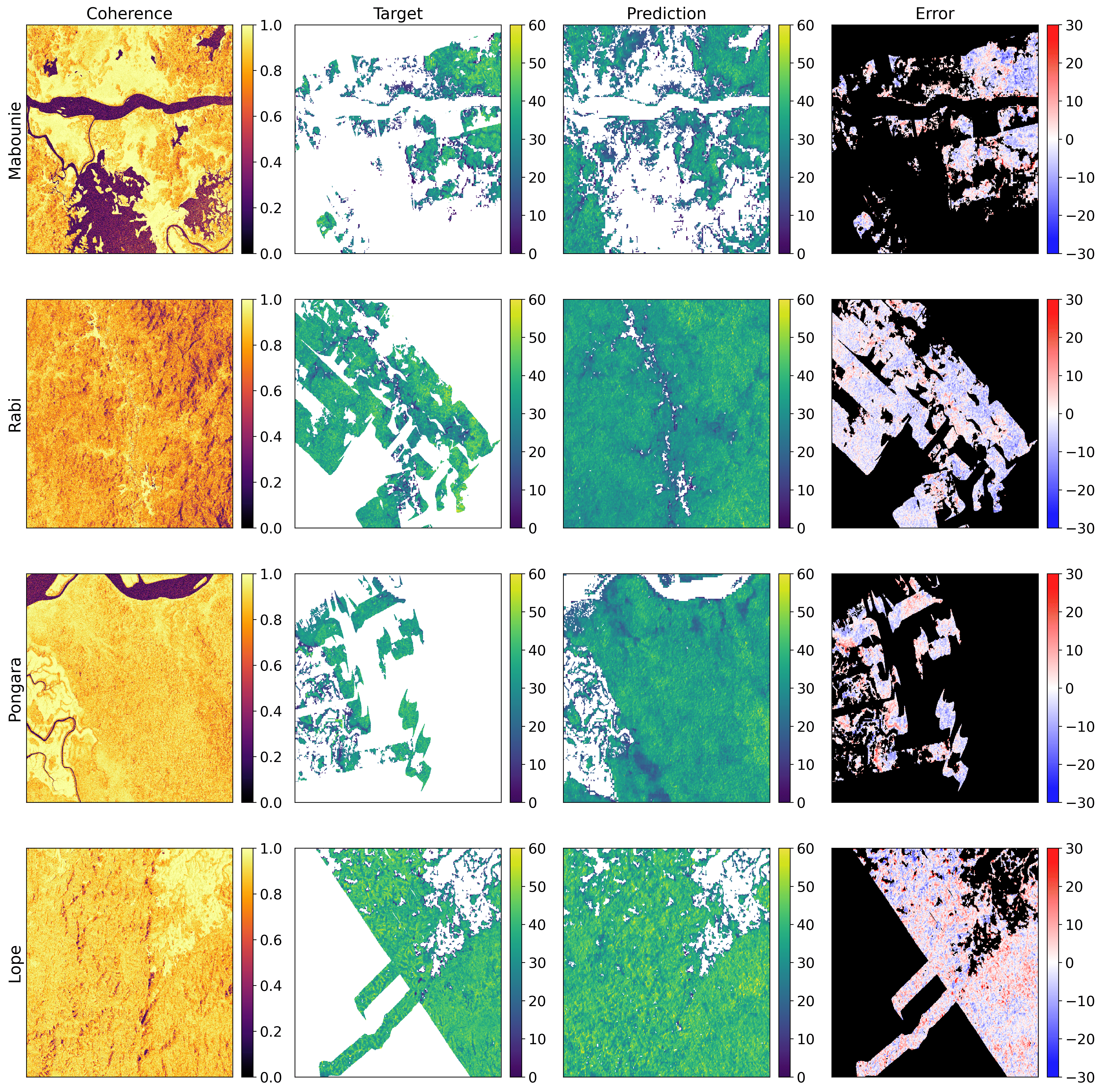}
    \caption{Qualitative results for CoHNet trained on Gabon over the four regions: Mabounie, Rabi, Pongara, and Lope. Areas with no available LVIS reference are shown in white. Predictions are masked by a forest/non-forest map (white regions denote areas with no forest).}
    \label{fig:feature_maps_all_regions}
\end{figure*}
\begin{figure*}[t]
    \centering
    \includegraphics[width=1\textwidth]{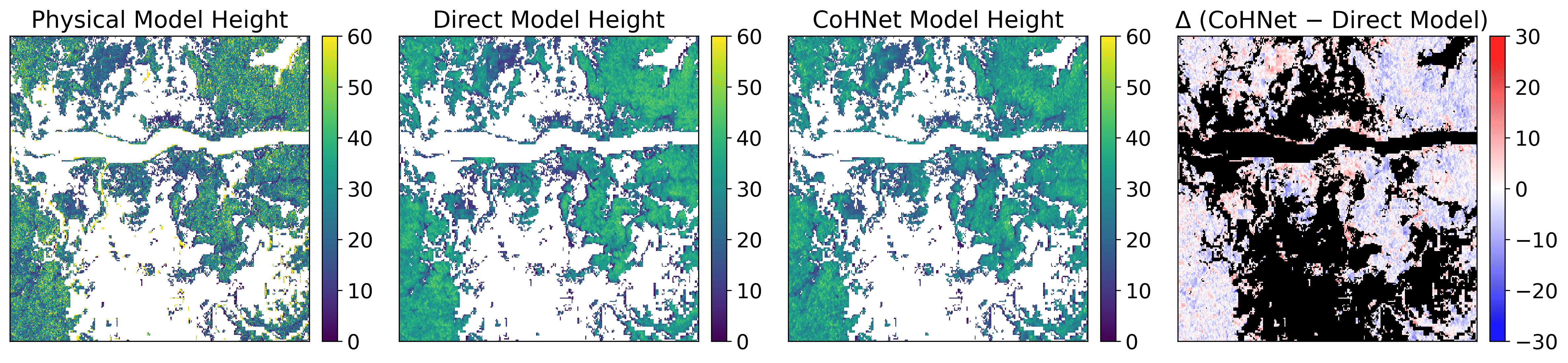}
    \caption{Qualitative comparison of forest height estimates from different models such as the physical model (RVoG), the direct model, CoHNet, and the difference (CoHNet - Direct Model) for the Mabounie region.}
    \label{fig:height_maps}
\end{figure*}

We first evaluate the performance of the proposed model over the different regions. 
We use the surrogate model trained over all of Gabon as it shows the best agreement with the physical model (see Section~\ref{sec:nsm}). 
The training of the surrogate model only requires input data and the height estimates of the physical model, which can be easily computed for all regions.
We leave the evaluation of the influence of different surrogate models to an ablation study shown in Section~\ref{sec:abl-nsm}.

The RMSE in Figure~\ref{fig:conf_matrix_End_to_End_Model_Gabon_PM} shows %excellent 
good performance of roughly 7~m RMSE if the model is trained and tested on the same region (note, that train and test data belong to different image acquisitions and different areas within the region). 
This decreases the RMSE by roughly 50\% compared to the traditionally used RVoG model (see Table~\ref{tab:stats}).
The error naturally increases if the model is applied to a region it is not trained on. 
However, in most of these cases the loss in performance is limited.
The best-performing model across all regions is the combined Gabon model for which the RMSE varies between 5.97 to 8.62~m. 

Performance of all the models drops when evaluated on the Lope region with the exception of the Lope model itself indicating a distribution shift in this region. 
The Rabi model struggles in other regions showing the highest errors with up to 11.3~m on Lope. 

Figure~\ref{fig:feature_maps_all_regions} shows qualitative results for the Mabounie region illustrating a good visual agreement between predictions and LVIS measurements.

\subsection{Comparison with a direct model}
To analyze the performance further, we compare against a U-Net model \cite{mahesh2024forest} that learns a direct input-output mapping between coherence and the reference LVIS forest heights without leveraging any physical constraints. 
Following the data setup described in Section~\ref{sec:data}, we train five independent models, i.e. a model for each single region and a combined model using data from all the four regions.

Figure~\ref{fig:conf_matrix_direct_model} shows the Gabon model to perform best across all regions with an RMSE between 5.2 to 6.71~m, i.e. a 4.2-12.1\% decrease compared to the region-specific models. 
This is in line with other forest height estimates over the region based on deep learning models, e.g. Carcereri et al. \cite{carcereri2023deep} reports an RMSE of 5.41~m (although not directly comparable due to differences in input and dataset splits).

The direct models have slightly lower errors in most cases with differences from 0.71 to 2.01 m. 
A performance decrease is noticed for the Rabi and Lope regions where CoHNet outperforms the direct model.
Figure~\ref{fig:height_maps} summarizes the different relevant height maps of the physical RVoG model, the direct model, and CoHNet (using the Gabon NSM).

\subsection{Ablation Study}
\label{sec:abl-nsm}
% We study the effectiveness of CoHNet by using the NSM of the respective region only, i.e. the CoHNet model trained on data of Mabounie also uses the NSM that is trained on Mabounie.
We study the effectiveness of CoHNet using each region's specific NSM, i.e. the CoHNet model trained on data of Mabounie also uses the NSM that is trained on Mabounie.

Figure~\ref{fig:conf_matrix_End_to_End_Model} %\textcolor{red}{TBD: In Figure 3 swap 3c with 3d so the order of reference is maintained.} 
shows similar RMSE values compared to the model with the Gabon NSM. There is an increase of error for Lope models where the generalization of the model decreases when evaluated on the other regions. Using a better physical model indicates that there is a possibility to reduce the error even further. The achieved performance critically depends on the definition of the inversion model. An oversimplified inversion model with strict assumptions results in a compromised performance of the model and the ability to generalize further. A more complex or more robust version of the RVoG model can help to resolve this issue and to improve the overall performance of the proposed approach. 

\subsection{Optimized Volume Decorrelation}
\begin{figure*}[t]
    \centering
    \includegraphics[width=0.95\textwidth]{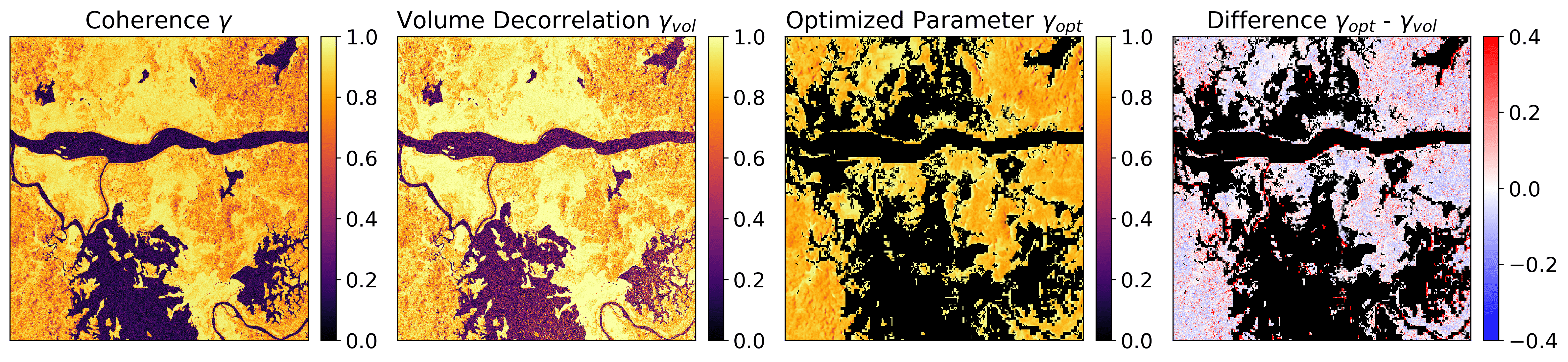}
    \caption{Qualitative comparison of different types of coherence features in CoHNet such as input coherence, volume decorrelation, optimized volume decorrelation, and the difference between the both for the Mabounie region. Black regions represent non-forest areas.}
    \label{fig:cohernece maps}
\end{figure*}

\begin{figure*}[t]
    \centering
    \begin{subfigure}{0.32\textwidth}
        \centering
        \includegraphics[width=\textwidth]{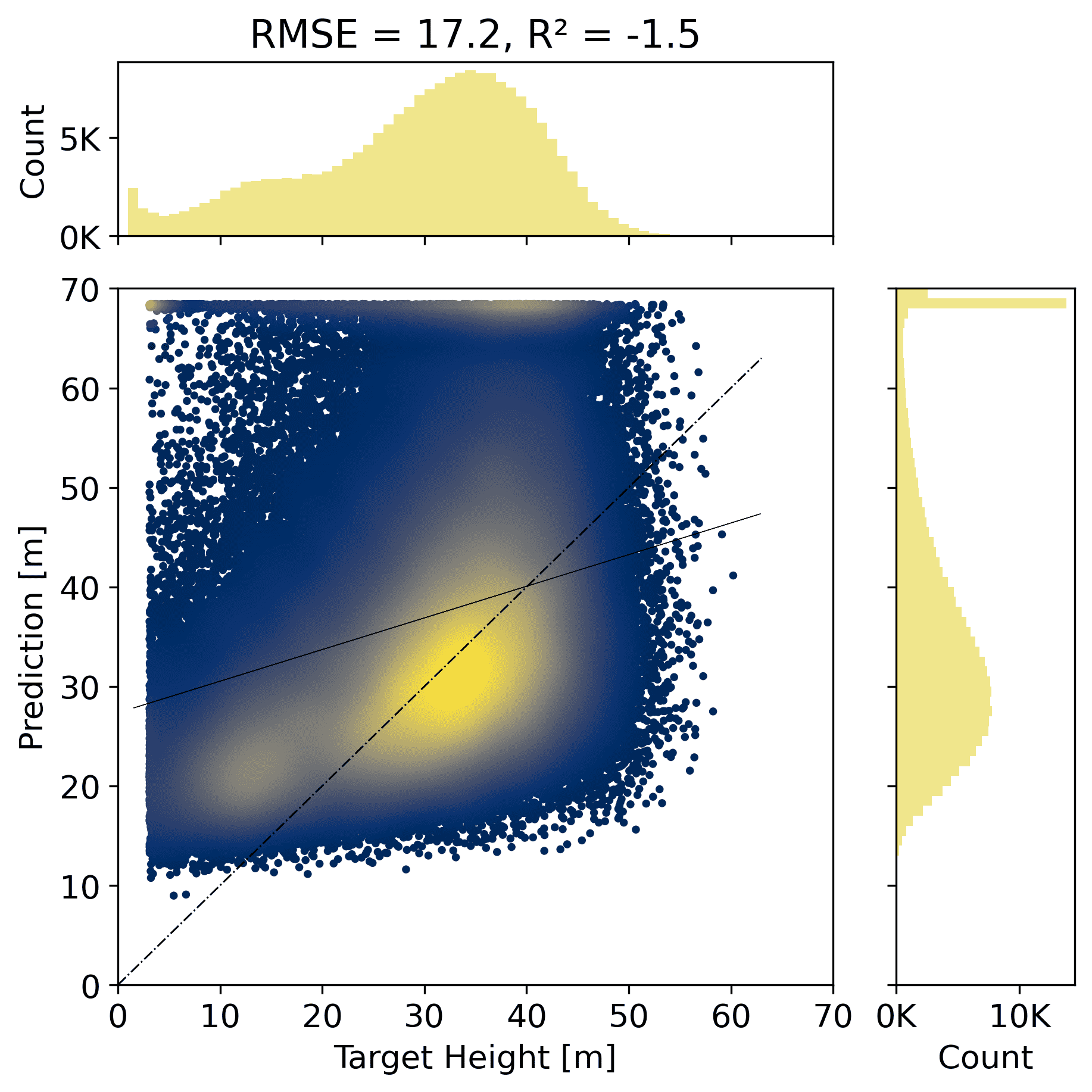}
        \caption{Coherence}
        \label{fig:Coherence Parameter}
    \end{subfigure}%
    \hspace{0.01\textwidth}
    \begin{subfigure}{0.32\textwidth}
        \centering
        \includegraphics[width=\textwidth]{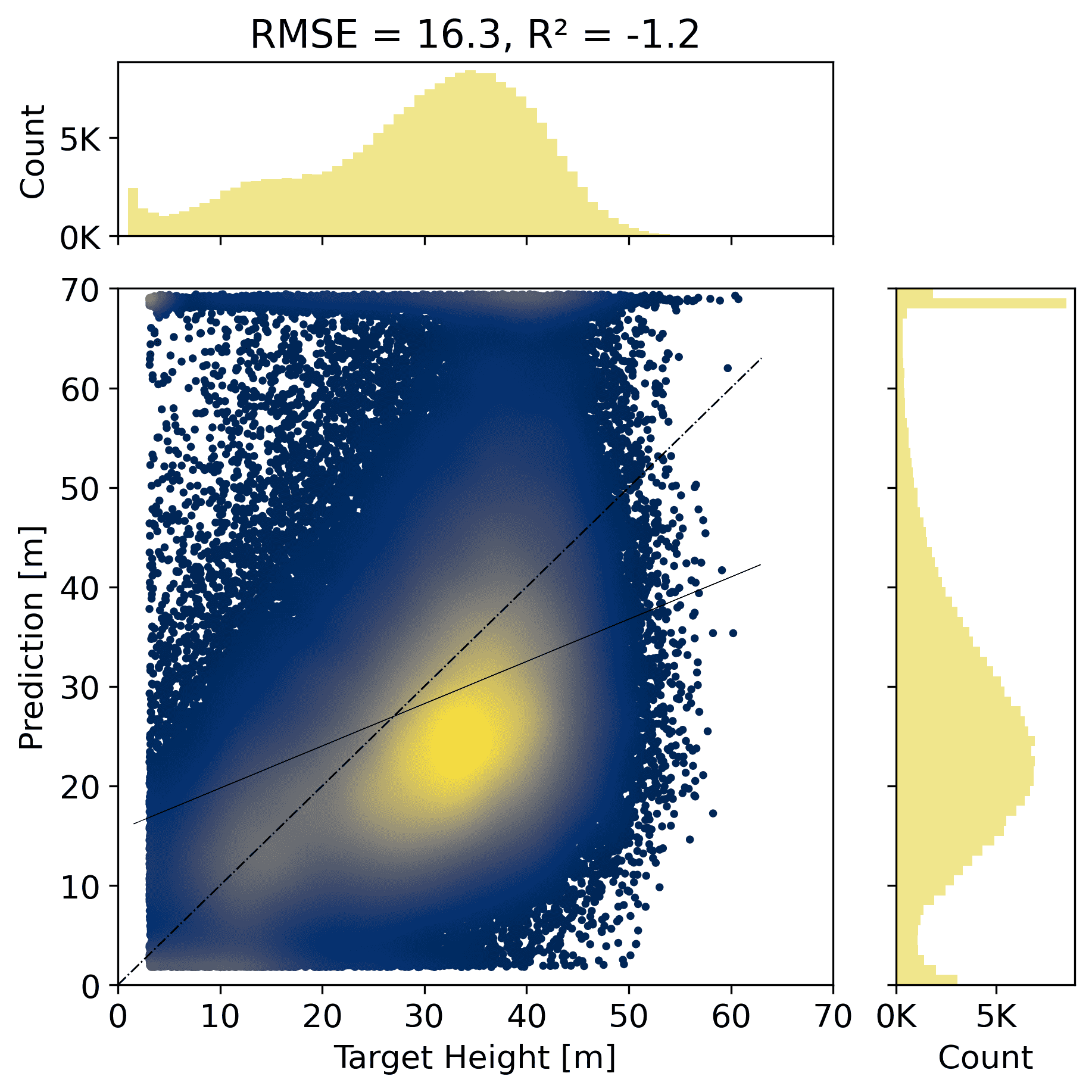}
        \caption{Volumetric Decorrelation}
        \label{fig:image2}
    \end{subfigure}%
    \hspace{0.01\textwidth}
    \begin{subfigure}{0.32\textwidth}
        \centering
        \includegraphics[width=\textwidth]{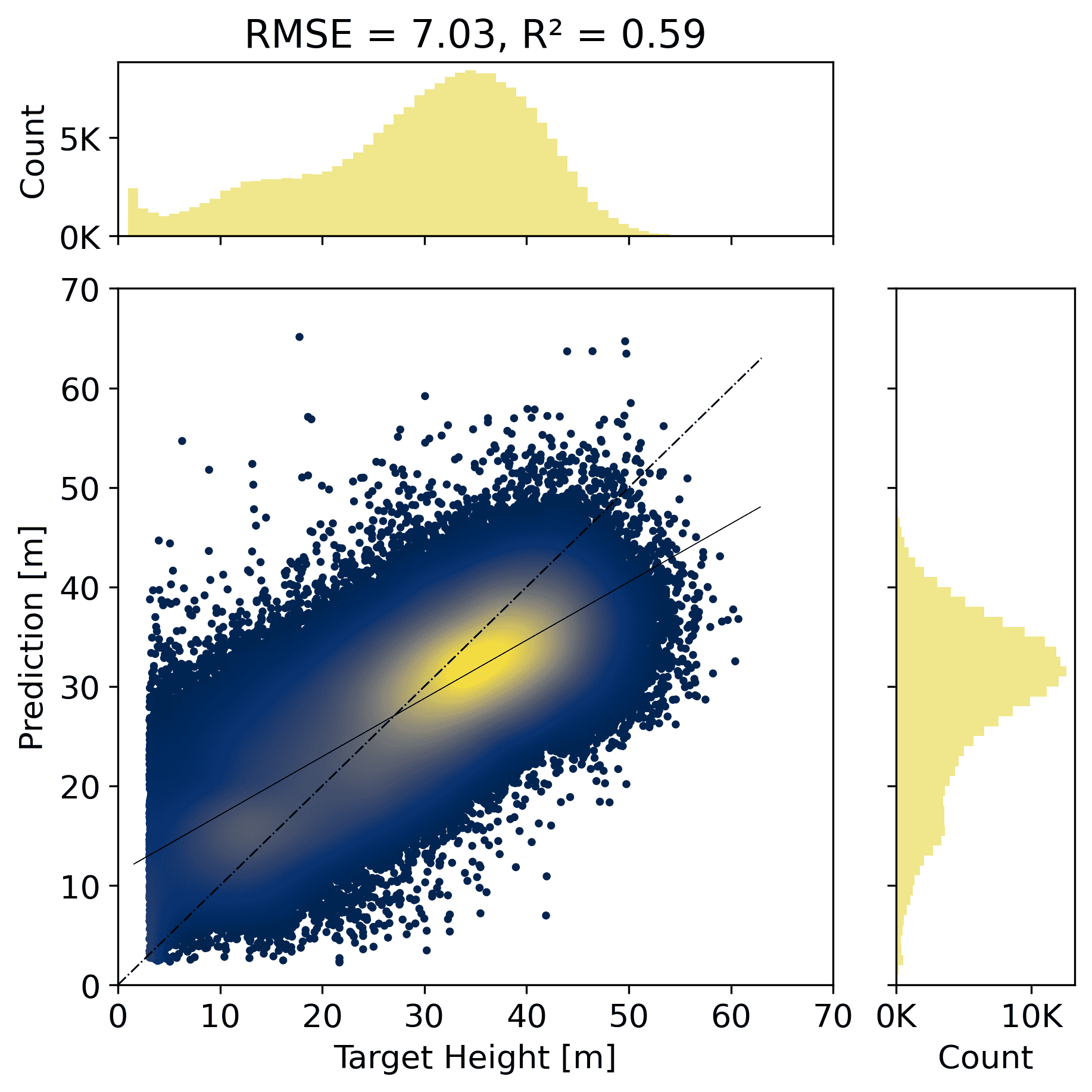}
        \caption{Optimized Volumetric Decorrelation}
        \label{fig:image3}
    \end{subfigure}
    %\\
    % \begin{subfigure}[b]{0.29\textwidth}
    %    \centering
    %    \includegraphics[width=\textwidth]{Images/density_scatter_Opt_parm_vs.Coh_13.png}
    %    \caption{Opt. Parameter vs. Coherence }
    %    \label{fig:pred}
    %\end{subfigure}%
    %\hspace{0.05\textwidth}
    %\begin{subfigure}{0.29\textwidth}
    %    \centering
    %    \includegraphics[width=\textwidth]{Images/density_scatter_Opt_parm_vs.Target_Vol_Decor_13.png}
    %    \caption{Opt. Parameter vs. Vol. Decorrlation}
    %    \label{fig:pred}
    %\end{subfigure}
    \caption{Density scatter plots for height inversion via a physical model for different input parameters.% and Optimized Parameter against Coherence and Volumetric Decorrlation [d, e].
    }
    \label{fig:scatter_plots}
\end{figure*}

The preceding sections show that CoHNet generates physically-consistent and accurate forest height.
However, in contrast to both the physical model and the direct model it offers a second output: The optimized estimate of the volume decorrelation which is of potential interest on its own and valuable also for other applications, e.g. being used in other forest height models.

Figure~\ref{fig:cohernece maps} compares the original coherence, the computed volume decorrelation, and the estimated optimized volume decorrelation over Mabounie. 
The latter is less noisy and provides finer structures. 
Figure~\ref{fig:scatter_plots} shows the results if each of these parameters is used for forest height inversion in a physical model. 
Using the original coherence overestimates small heights, while heights around 30-40~m are underestimated yielding an RMSE of 17.2~m.
The overestimation of smaller heights is well compensated by using the computed volume decorrelation, reducing the overall RMSE to 16.3~m.
Finally, leveraging the estimated optimized volume decorrelation improves the estimation for all height levels and reduces many of the existing ambiguities, decreasing the RMSE to 7.03~m.
\section{Conclusion}
We introduce CoHNet - an end-to-end, physics-informed model, which generates physically plausible forest height estimates. CoHNet uses a U-Net backbone in a pipeline where two models are concatenated: The first estimates optimized volume decorrelation while the second uses a physical model to predict forest height. %We introduce a novel physics based training loss.
%We demonstrate the ability of CoHNet to estimate optimized volume decorrelation and forest height. 
CoHNet leverages input data such as coherence and reference forest height, but no reference data for volume decorrelation is needed.

The neural surrogate model requires additional training compared to the direct model. However, the corresponding training data consists only of data products that are readily available, i.e. mainly coherence and the forest height computed by the physical model. These data products can easily be obtained for all regions where InSAR data is available and the physical model is applicable. 

CoHNet reduces the error of the physical model by roughly 50\% and is on par with a direct deep learning model while maintaining the advantages of a PM and providing optimized volume decorrelation as additional output - a variable that is of relevance in a manifold of applications.
% and compares favorably against the direct model while providing physically meaningful insights via constraining forest height with a physical model and estimating optimized volume decorrelation - a variable that is of relevance in a manifold of applications.
%While maintaining the advantages of a PM, CoHNet offers an accuracy which is on par with a direct deep learning model and provides optimized volume decorrelation as additional output.
Volume decorrelation is a key InSAR feature when dealing with vegetation and fundamental in applications that rely on estimating structural properties of volumetric targets.
In addition to estimating forest height, it enables biomass estimation which is important for carbon stock assessments and ecological studies. So far, volume decorrelation is derived from InSAR coherence via empirical heuristics. CoHNet is the first data-driven approach which does not apply hard-coded rules but estimates it directly from coherence.

A limitation of CoHNet is that the derived optimized volume decorrelation is only meaningful over forest. As it is directly optimized through matching the measured forest height, there are no constraints applied for regions that do not contain forest - or more specifically where no reference height is available and where the physical model is not applicable. 
A disadvantage of the used dataset is the fixed resolution of 20~m which may not capture fine-scale forest structure variation.  This can addressed by using high-resolution LiDAR and SAR measurements. 
%It can be extended to include other tropical rainforests (e.g. Amazon) to create a global model for tropical rainforests.

Future work will focus on integrating a detection module predicting when the physical model is applicable. %can be applied or whether the model should fall back to a direct solution. 
Furthermore, the uncertainty of the surrogate model in matching the given physical model can be taken into account to further enhance accuracy and robustness. %\textcolor{red}{A possible future extention of our work could include exploring Transformer-based architectures to capture global-contextual information and validate hypothesis for improved accuracy and generalization.}

\FloatBarrier 
{
    \small
    \bibliographystyle{ieeenat_fullname}
    \bibliography{main}
}

\clearpage
\setcounter{page}{1}
\maketitlesupplementary

\section{Random Volume over Ground Model}
\label{sec:rvog}
\begin{figure}[ht]
    \centering
    % \includesvg[width=1\linewidth]{Images/RVOG_fig.svg}
    \includegraphics[width=1\linewidth]{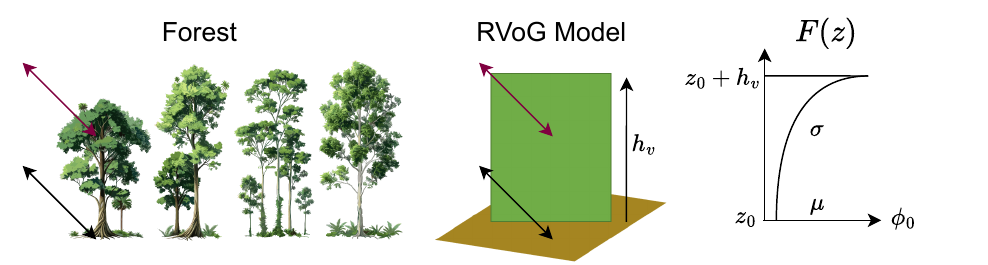}
    \caption{The RVoG model describes the forest structure as a two-layer scattering model with ground elevation~$z_0$ and volume height~$h_v$. $F(z)$ is the radar reflectivity of the forest scatters at different heights~$z$ and decays as a function of the extinction coefficient $\sigma$, the ground phase~$\phi_0$ and the ground-to-volume amplitude ratio~$\mu$.}
    \label{fig:rvog}
\end{figure}

The RVoG model illustrated in Figure~\ref{fig:rvog} is one of the most commonly used physcial models for InSAR-based forest height inversions. 

\section{Study Sites of the AfriSAR Campaign}
\label{sec:sites}
% \begin{figure}[ht]
%     \centering
%     \includegraphics[width= 1\linewidth]{Images/Dataset Location.png}
%     \caption{Overview of the dataset region over Gabon including the test regions Mabounie, Rabi, Pongara, and Lope }
%     \label{fig:dataset_loc}
% \end{figure}

\begin{figure*}[t]
    \centering
    \begin{subfigure}{0.45\textwidth}
        \centering
        \includegraphics[width=\textwidth]{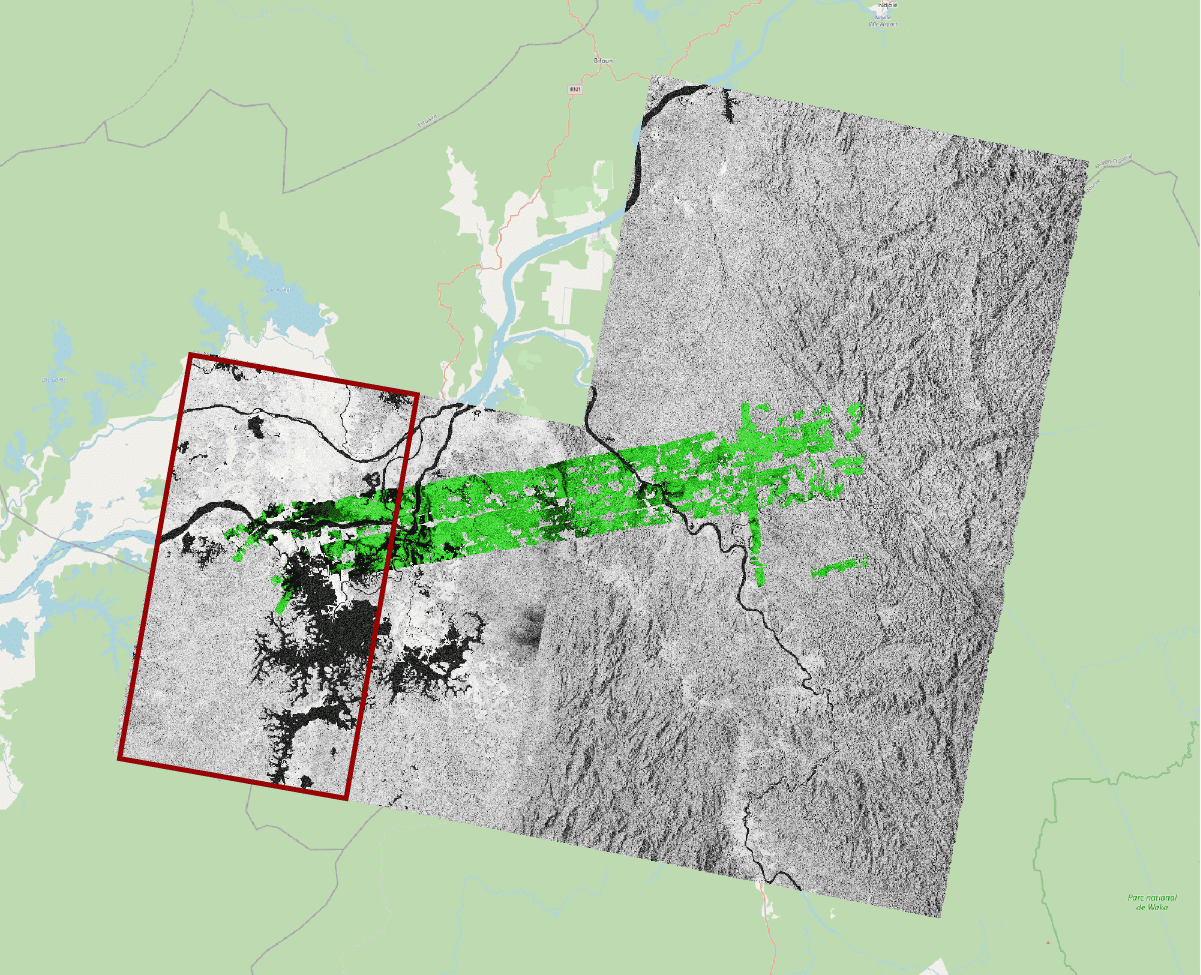}
        \caption{Mabounie}
        \label{fig:mabo_region}
    \end{subfigure}%
    \hspace{0.02\textwidth}
    \begin{subfigure}{0.45\textwidth}
        \centering
        \includegraphics[width=\textwidth]{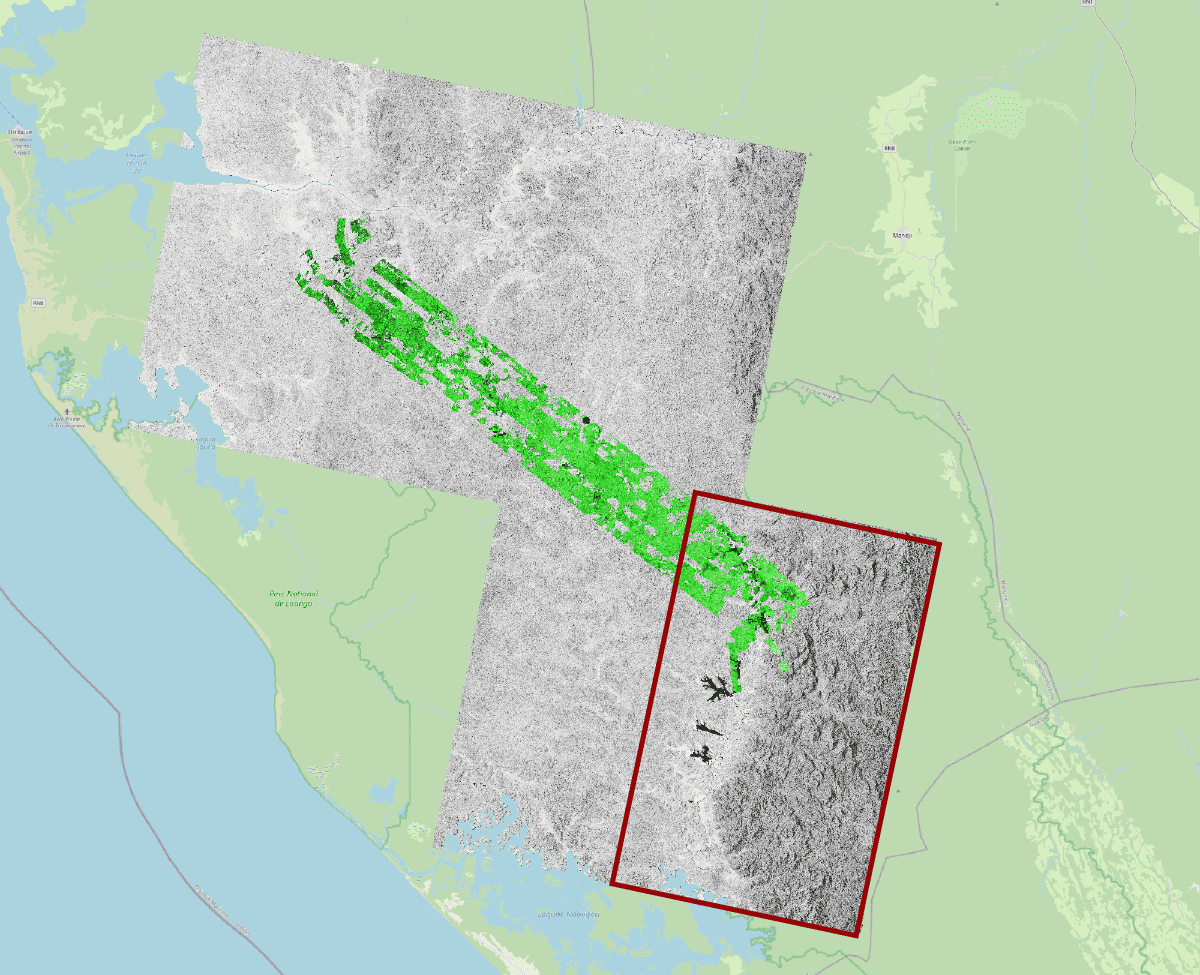}
        \caption{Rabi}
        \label{fig:rabi_region}
    \end{subfigure}
    \\
    \begin{subfigure}{0.45\textwidth}
        \centering
        \includegraphics[width=\textwidth]{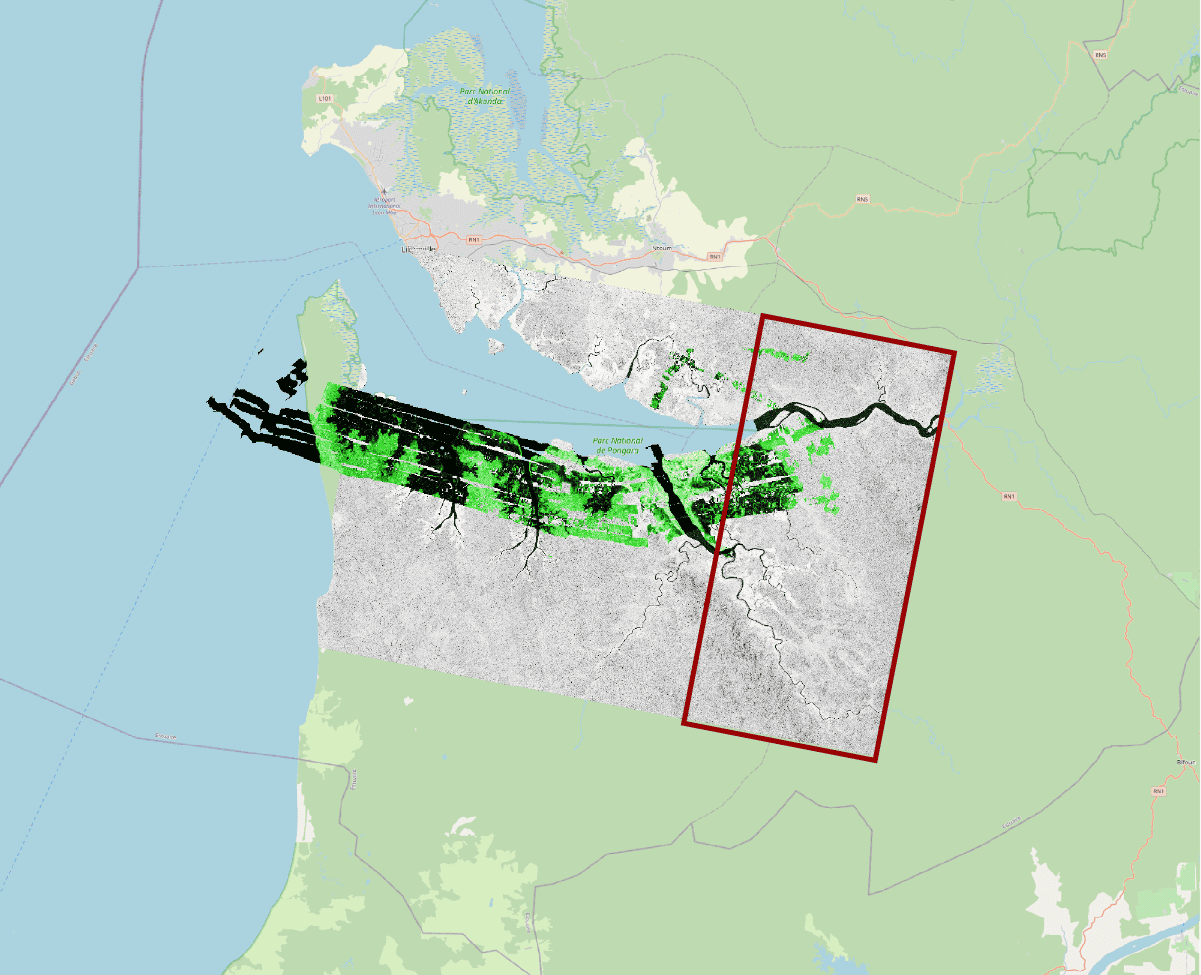}
        \caption{Pongara}
        \label{fig:pong_region}
    \end{subfigure}%
    \hspace{0.02\textwidth}
    \begin{subfigure}{0.45\textwidth}
        \centering
        \includegraphics[width=\textwidth]{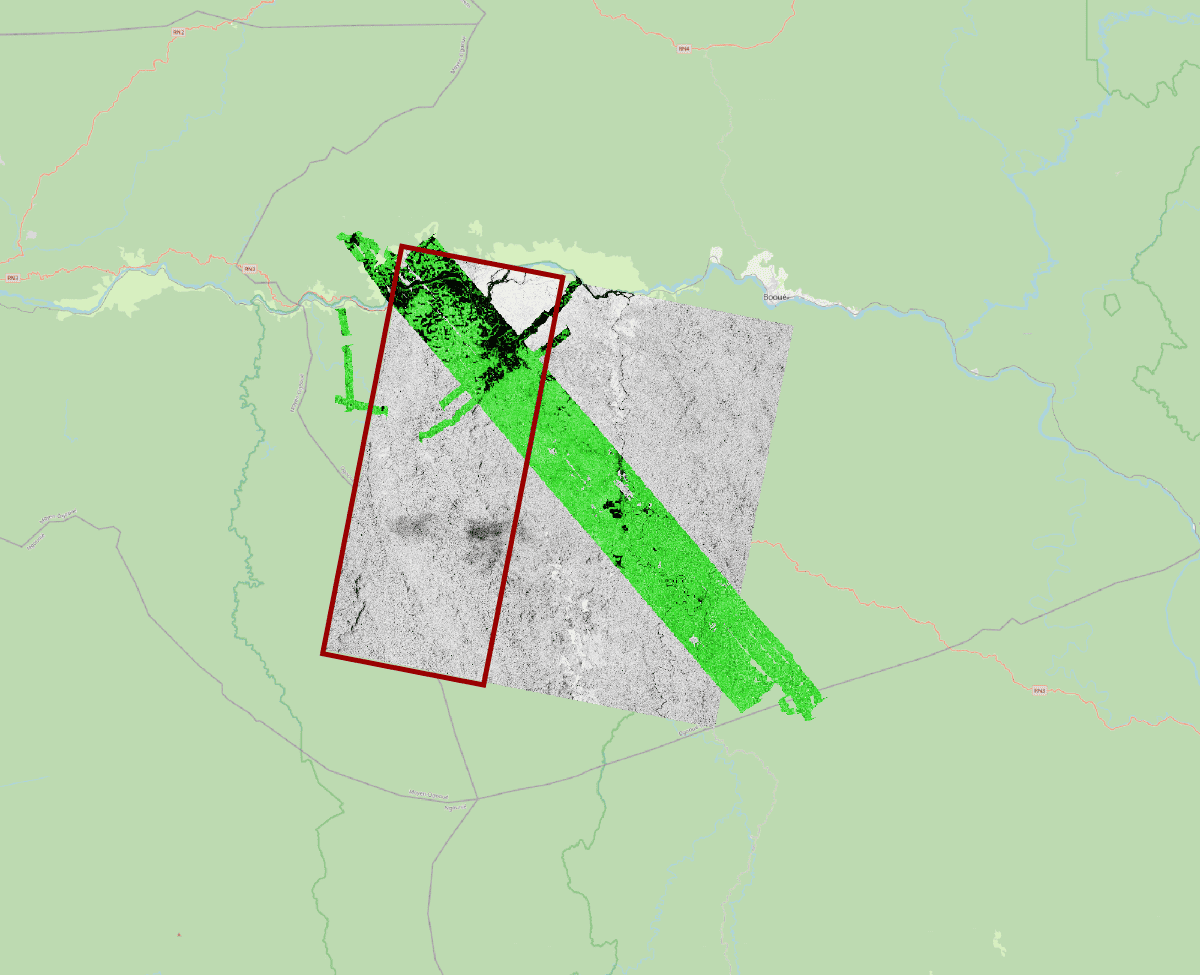}
        \caption{Lope}
        \label{fig:lope_region}
    \end{subfigure}%
    \caption{Dataset visualization with training and test areas for the used regions of Mabounie, Rabi, Lope, and Pongara. The bright green color represents the target height measurements of LVIS overlaid on top of TanDEM-X coherence. Red rectangles indicate the test areas.}
    \label{fig:all_regions}
\end{figure*}

During training, reference height from the AfriSAR campaign is used. The study area for the campaign is Gabon as it is densely forested country with rich structural and functional biodiversity. During the AfriSAR campaign, over 7000 $km^{2}$ of waveform Lidar data from LVIS were collected from 10 key sites. The four regions used for training and testing are shown in Figure~\ref{fig:all_regions} and geographical coordinates are given in Table~\ref{tab:coordinates}. The target data used in our study is open-access through AfriSAR campaign and avaialble via Nasa's EarthData portal.

The corresponding SAR acquisitions for this region are obtained by the TanDEM-X (DLR) satellite mission. In 2016, TanDEM-X was operated on the coinciding regions of the AfriSAR region for further scientific experiments with bistatic modes of operation. The SAR images used in this study are the CoSSC (Co-registered Single Look Slant Range Complex) product and derived features. Bistatic sensors such as TanDEM-X consist of two satellites flying in a close formation separated by a spatial baseline acquiring SAR data simultaneously from the same area. The two acquired SAR images are focused and spatially aligned before being used for further processing. The resulting dataset after performing computations for coherence, volumetric decorrelation, and resampling the LVIS Heights from AfriSAR campaign has a  storage size of $\sim$19GB.
% The InSAR-derived feature complex coherence is the pixel-wise product of the SLCs (with one of them complex conjugated) and normalized with both their magnitudes.  It is a cross-correlation factor providing insights into the coherence [absolute value range 0 - 1] or similarity of radar scattering properties between two acquisitions. High coherence values are consistent with stable scatterers (eg. building) present in the scene and low coherence values are indicative of a change in scatterers due to various decorrelation factors and contain valuable information about volumetric scattering due to vegetation

\begin{table*}[b] %!ht
\centering
\begin{tabular}{|l|l|l|l|l|}
\hline
\textbf{Site} & \textbf{Westernmost Longitude} & \textbf{Easternmost Longitude} & \textbf{Northernmost Latitude} & \textbf{Southernmost Latitude} \\ \hline
Mabounie & 9.935094 & 10.7425 & -0.67959 & -0.95097 \\ \hline
Rabi & 9.594942 & 10.27658 & -1.68464 & -2.29494 \\ \hline
Pongara & 9.179286 & 9.995917 & 0.265536 & 0.004522 \\ \hline
Lope & 11.39674 & 12.02394 & -0.01559 & -0.64224 \\ \hline
\end{tabular}
\caption{Geographical Coordinates of regions from Gabon used in this study.}
\label{tab:coordinates}
\end{table*}

\section{Performance Metrics}
\label{sec:metrics}
To test model performance we use the Root Mean Squared Error (RMSE) 
\begin{equation}
\text{RMSE} = \sqrt{\frac{1}{n} \sum_{i=1}^{n} (y_i - \hat{y}_i)^2}
\label{eq:rmse}
\end{equation}
and the coefficient of determination $R^2$
\begin{equation}
R^2 = 1 - \frac{\sum_{i=1}^{n} (y_i - \hat{y}_i)^2}{\sum_{i=1}^{n} (y_i - \bar{y})^2},
\label{eq:r2}
\end{equation}
where $y_i$ is the reference height measurement (and $\bar{y}$ its mean value) and $\hat{y}_i$ is the predicted height. 

\section{Further CoHNet Qualitative Results}
\label{sec:results}

\begin{figure*}[t]
    \centering
    \begin{subfigure}{0.24\textwidth}
        \centering
        \includegraphics[width=\textwidth]{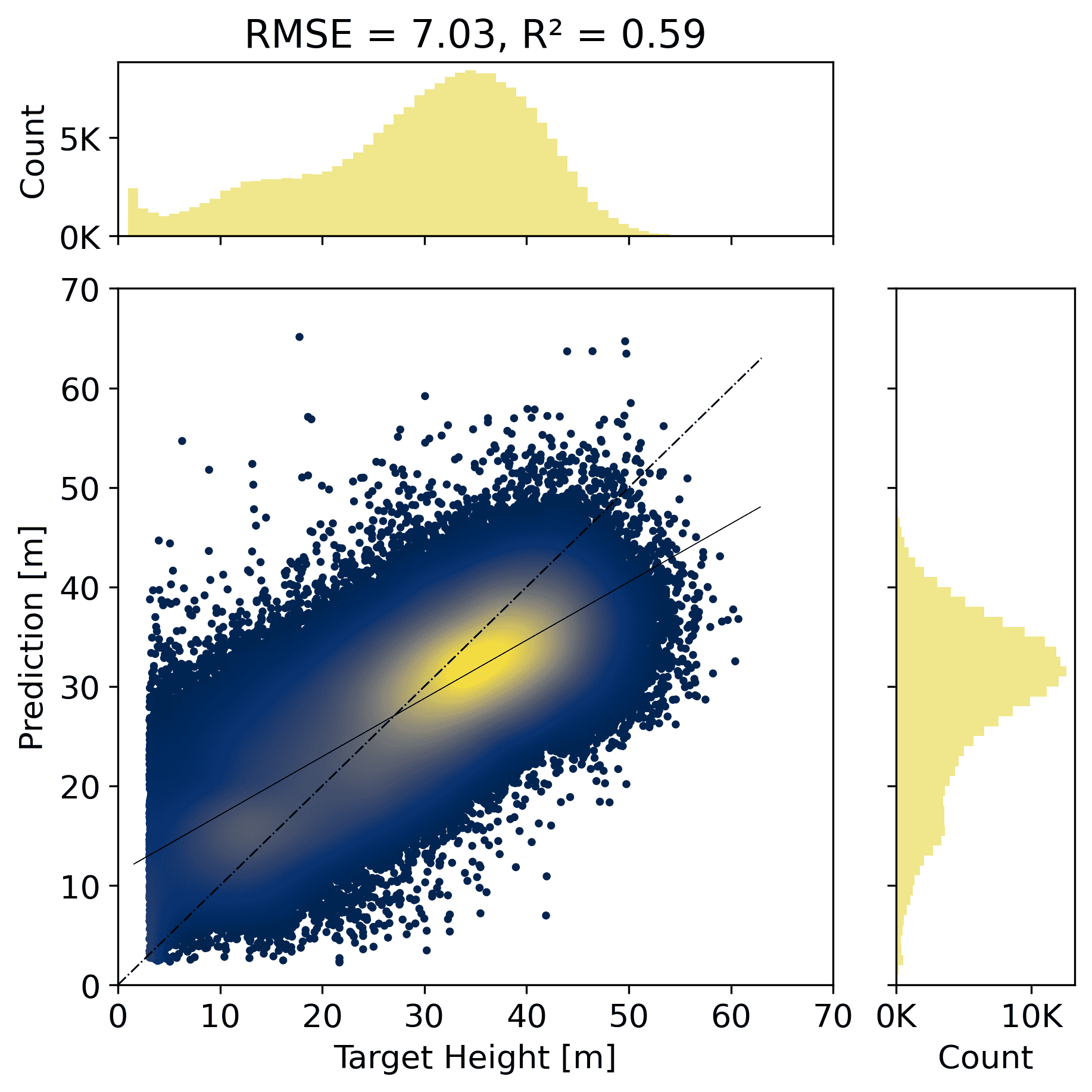}
        \caption{Mabounie}
        \label{fig:dens_mabo_region}
    \end{subfigure}%
    %\hspace{0.02\textwidth}
    \begin{subfigure}{0.24\textwidth}
        \centering
        \includegraphics[width=\textwidth]{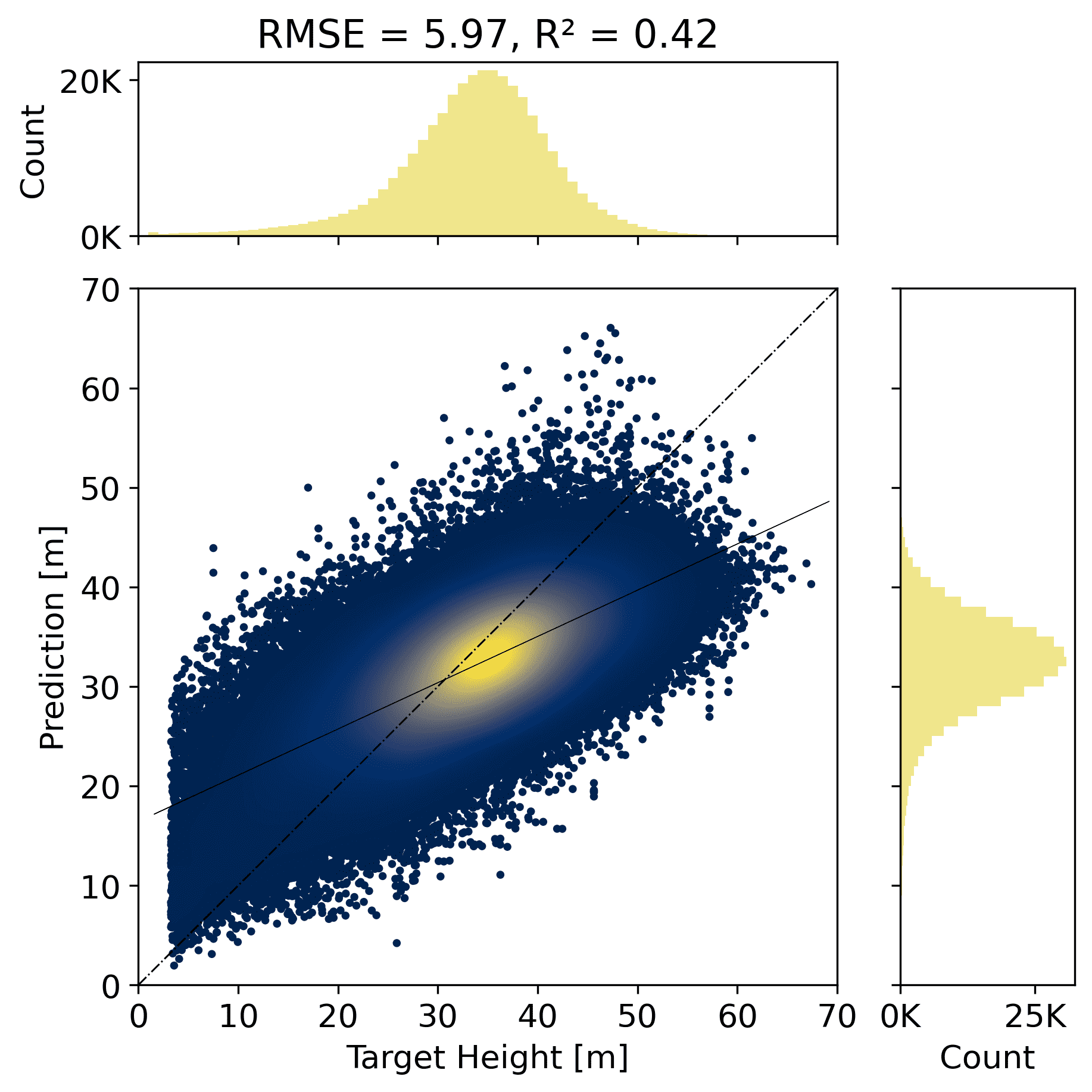}
        \caption{Rabi}
        \label{fig:dens_rabi_region}
    \end{subfigure}
    %\hspace{0.02\textwidth}
    \begin{subfigure}{0.24\textwidth}
        \centering
        \includegraphics[width=\textwidth]{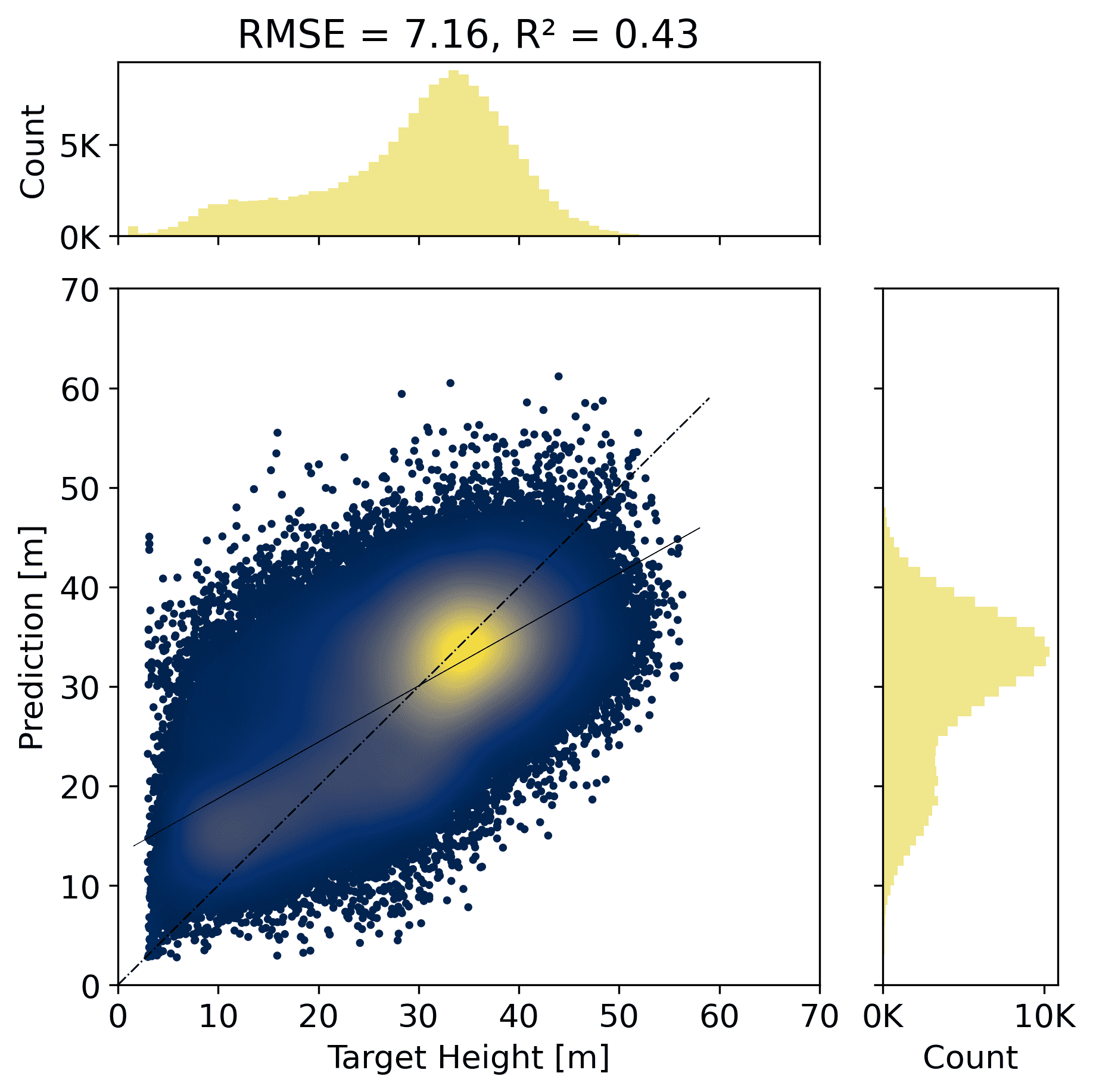}
        \caption{Pongara}
        \label{fig:dens_pong_region}
    \end{subfigure}%
    %\hspace{0.02\textwidth}
    \begin{subfigure}{0.24\textwidth}
        \centering
        \includegraphics[width=\textwidth]{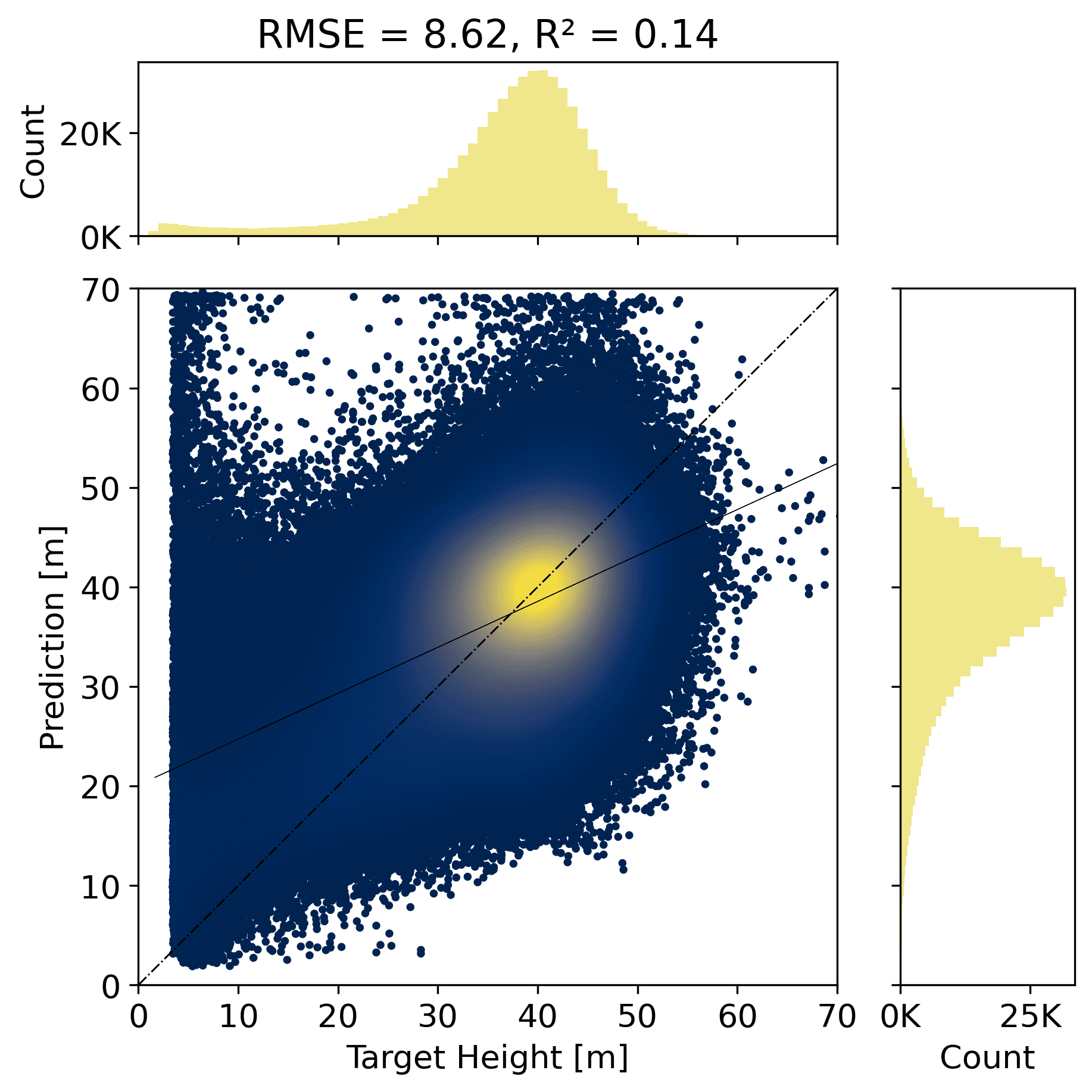}
        \caption{Lope}
        \label{fig:dens_lope_region}
    \end{subfigure}%
    \caption{Density scatter plots for height inversions with neural surrogate model using optimized volumetric decorrelation from CoHNet.}
    \label{fig:all_density}
\end{figure*}

\begin{figure*}
    \centering
    \includegraphics[width=1\textwidth]{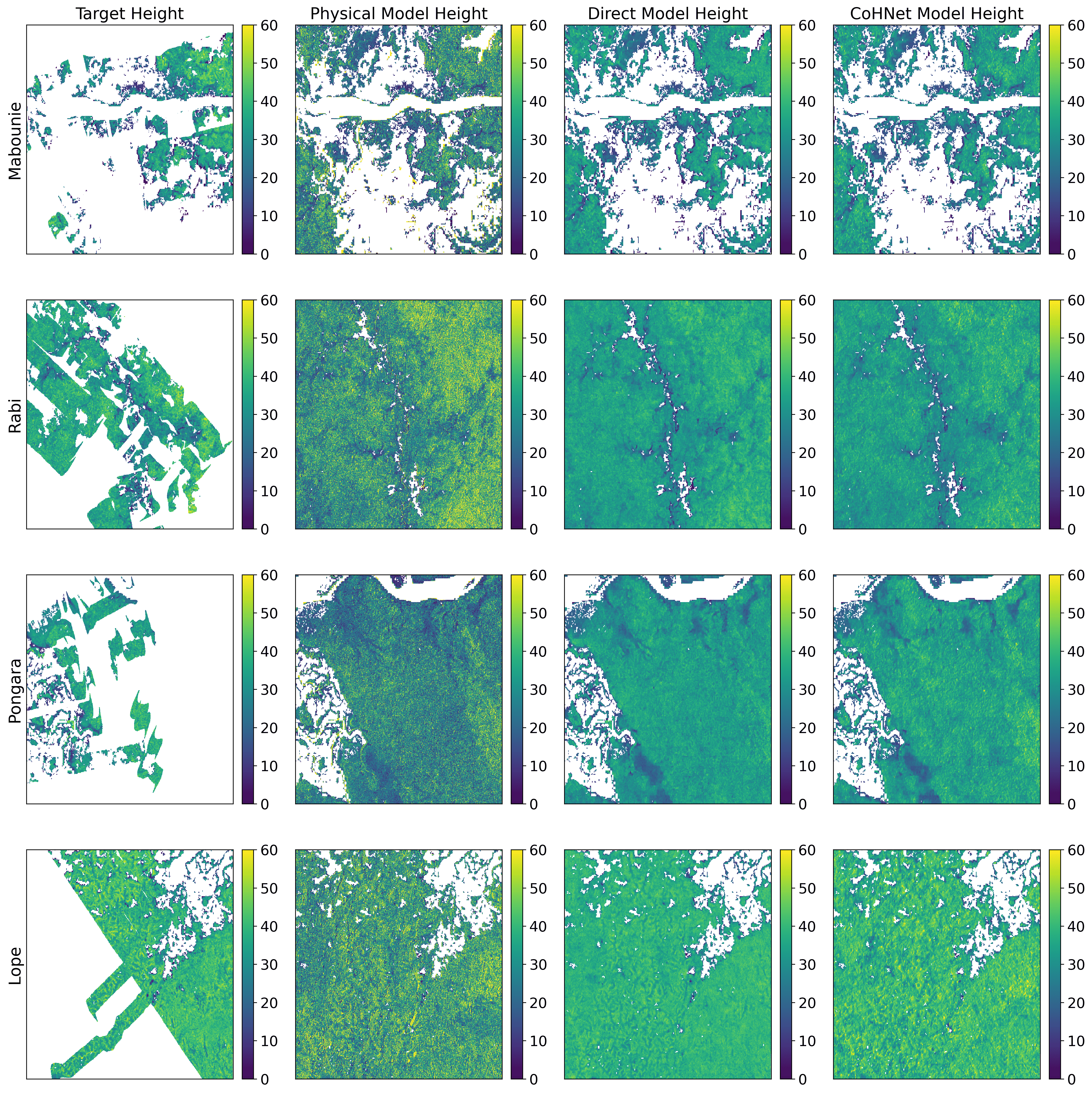}
    \caption{From left to right: Forest height as provided as reference value by LVIS and estimated by the physical model (RVoG), the direct model, and CoHNet for all used regions using the Gabon model. White regions in the reference data denote lack of LVIS measurements, while white regions in the model predictions denote non-forest areas for which the models are not applicable.}
    \label{fig:all_height_maps}
\end{figure*}

\begin{figure*}
    \centering
    \includegraphics[width=1\textwidth]{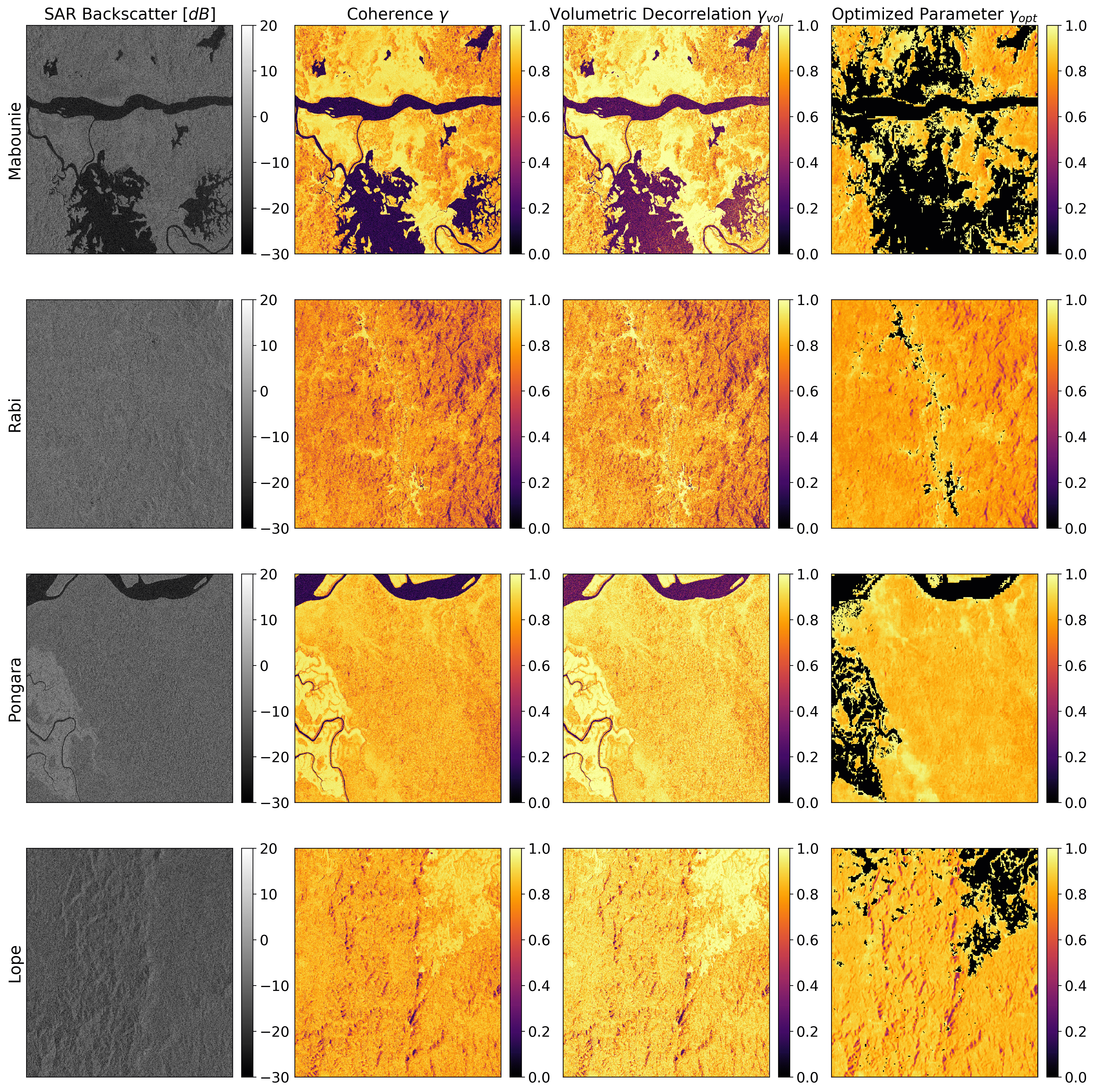}
    \caption{From left to right: SAR backscatter, input coherence $\gamma$, volumetric decorrelation~$\gamma_\text{Vol}$, and optimized volume decorrlation $\hat\gamma_\text{Vol}$ for all used regions. Black regions in the maps denote non-forest areas.}
    \label{fig:all_coh_maps}
\end{figure*}

Figure~\ref{fig:all_density} shows further quantitative results using density scatter plots with marginal height distributions for CoHNet forest estimates. While Figure~\ref{fig:all_height_maps} and Figure~\ref{fig:all_coh_maps} show qualitative results of forest height prediction and estimated coherence values over all used Gabon regions such as Mabounie, Rabi, Pongara, and Lope.

\begin{figure*}
    \centering
    \includegraphics[width=1\textwidth]{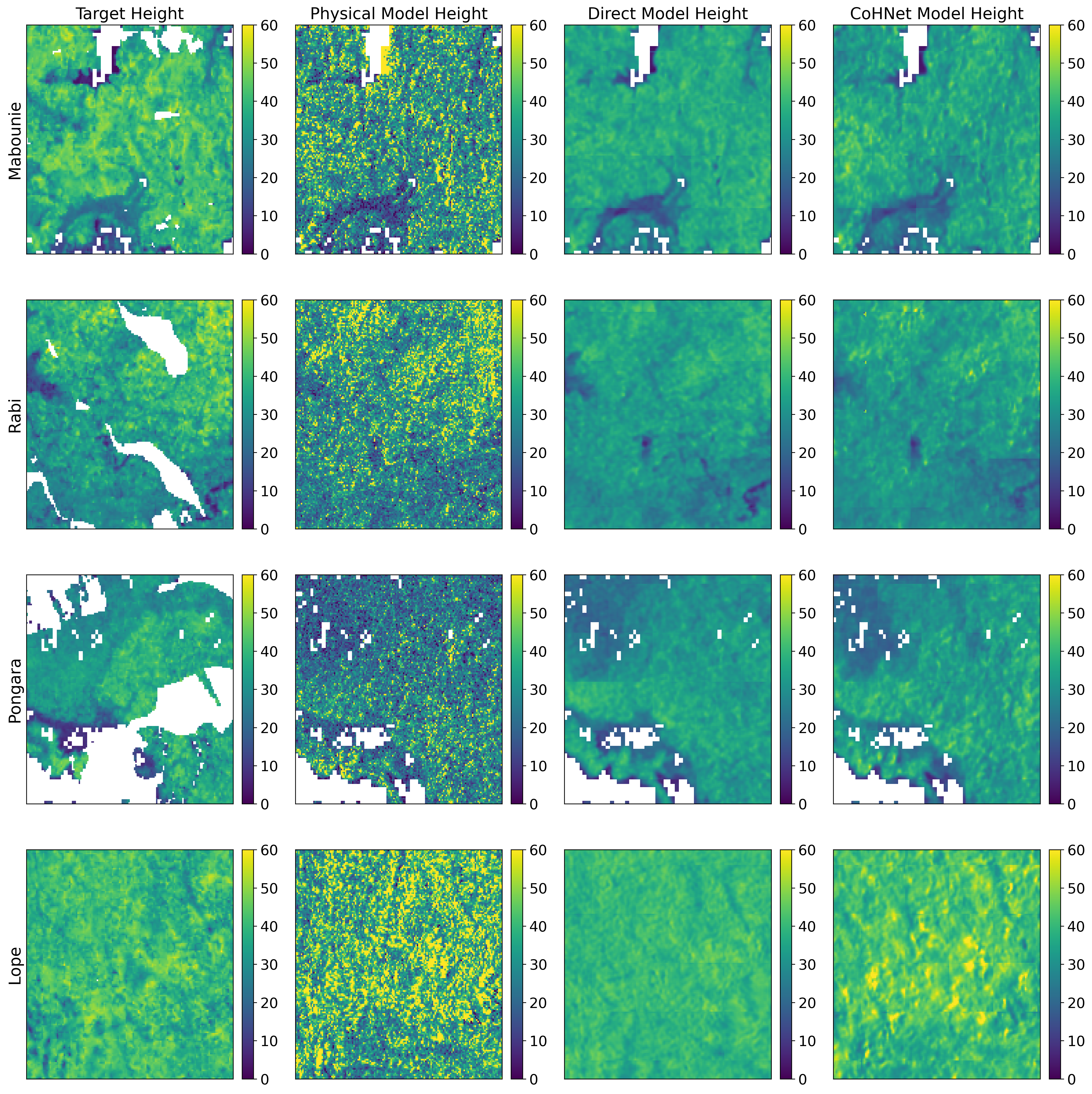}
    \caption{Zoomed in comparison, from left to right: Forest height as provided as reference value by LVIS and estimated by the physical model (RVoG), the direct model, and CoHNet for all used regions using the Gabon model. White regions in the reference data denote lack of LVIS measurements, while white regions in the model predictions denote non-forest areas for which the models are not applicable.}
    \label{fig:all_height_maps_zoom}
\end{figure*}

\begin{figure*}
    \centering
    \includegraphics[width=1\textwidth]{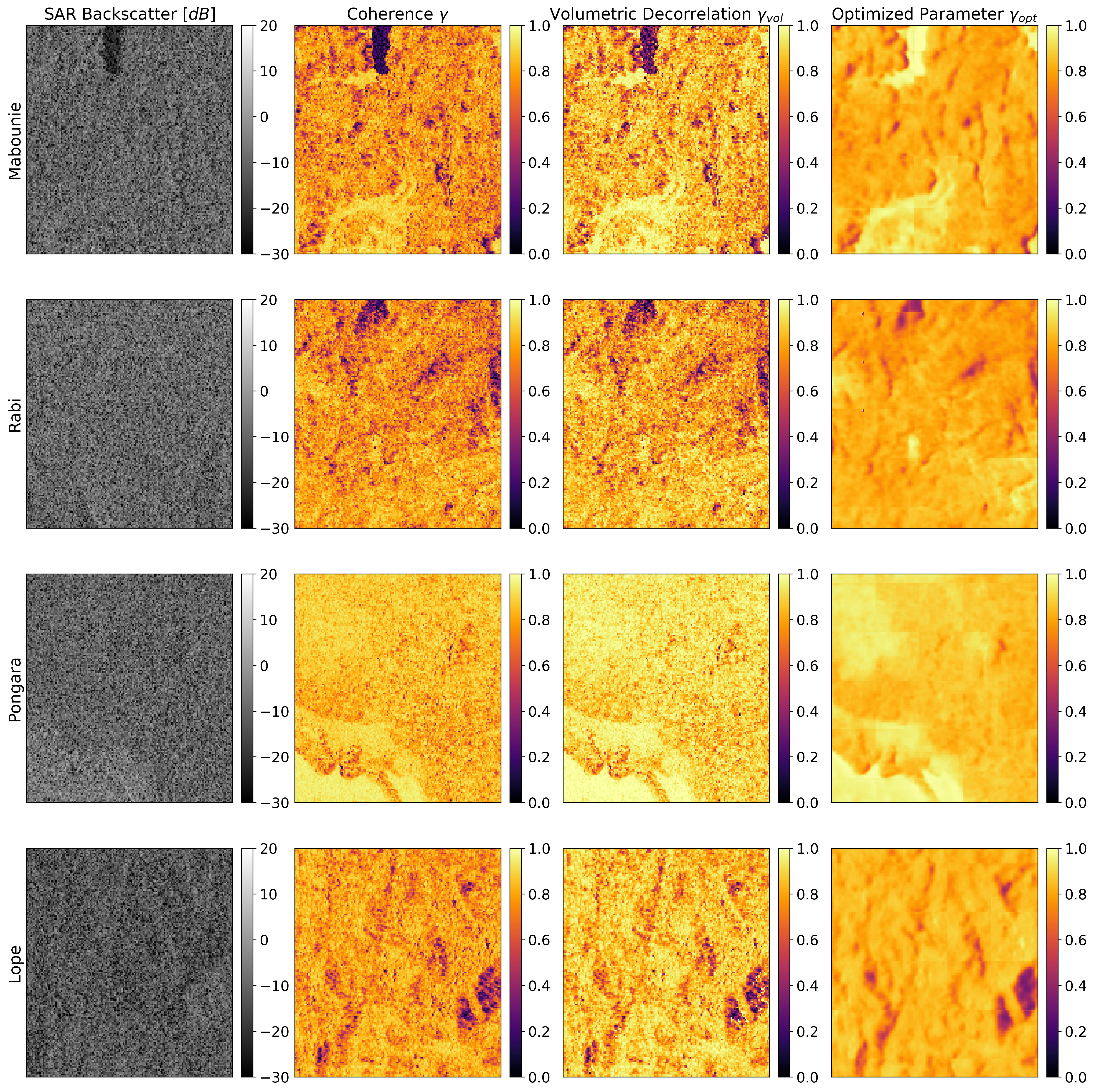}
    \caption{Zoomed in comparison, from left to right: SAR backscatter, input coherence $\gamma$, volumetric decorrelation~$\gamma_\text{Vol}$, and optimized volume decorrlation $\hat\gamma_\text{Vol}$ for all used regions.}
    \label{fig:all_coh_maps_zoom}
\end{figure*}

% \begin{figure*}[ht]
% \centering
% \begin{subfigure}[b]{0.29\textwidth}
%         \centering
%         \includegraphics[width=\textwidth]{Images/density_scatter_Opt_parm_vs.Coh_13.png}
%         \caption{Opt. Parameter vs. Coherence }
%         \label{fig:pred}
%     \end{subfigure}%
%     \hspace{0.05\textwidth}
%     \begin{subfigure}{0.29\textwidth}
%         \centering
%         \includegraphics[width=\textwidth]{Images/density_scatter_Opt_parm_vs.Target_Vol_Decor_13.png}
%         \caption{Opt. Parameter vs. Vol. Decorrlation}
%         \label{fig:pred}
%     \end{subfigure}
%     \caption{Density scatter plots for Optimized Parameter against Coherence and Volumetric Decorrlation .}
%     \label{fig:scatter_plots}
% \end{figure*}

%To split the supplementary pages from the main paper, you can use \href{https://support.apple.com/en-ca/guide/preview/prvw11793/mac#:~:text=Delete%20a%20page%20from%20a,or%20choose%20Edit%20%3E%20Delete).}{Preview (on macOS)}, \href{https://www.adobe.com/acrobat/how-to/delete-pages-from-pdf.html#:~:text=Choose%20%E2%80%9CTools%E2%80%9D%20%3E%20%E2%80%9COrganize,or%20pages%20from%20the%20file.}{Adobe Acrobat} (on all OSs), as well as \href{https://superuser.com/questions/517986/is-it-possible-to-delete-some-pages-of-a-pdf-document}{command line tools}.

\end{document}